\documentclass[10pt,journal,compsoc]{IEEEtran}
\usepackage{booktabs} 
\usepackage{xcolor}
\usepackage{soul}
\usepackage{epsfig}
\usepackage{graphicx}
\usepackage{amsmath}
\usepackage{amssymb}
\usepackage{courier}
\usepackage{helvet}
\usepackage{courier}
\usepackage{algorithm}
\usepackage{algorithmic}
\usepackage{multirow}
\usepackage{url}
\usepackage{array}
\usepackage{paralist}
\usepackage{balance}

\newcommand{\x}{{\bf x}}
\newcommand{\w}{{\bf w}}

\newcommand{\vvv}{{\bf v}}

\newcommand{\G}{\mathcal{G}}

\newcommand{\E}{\mathcal{E}}
\newcommand{\X}{\mathcal{X}}
\newcommand{\R}{\mathcal{R}}

\newcommand{\EE}{\mathbb{E}}

\newtheorem{defn}{Definition}

\ifCLASSOPTIONcompsoc
  \usepackage[nocompress]{cite}
\else
  \usepackage{cite}
\fi
\ifCLASSINFOpdf
\else
\fi
\begin{document}
%
\title{Semi-Supervised Image Captioning Considering Wasserstein Graph Matching}
%
%
%
%

\author{Yang Yang~\IEEEmembership{}
	\IEEEcompsocitemizethanks{\IEEEcompsocthanksitem Yang Yang is with the Nanjing University of Science and Technology, Nanjing 210094, China.\protect\\
		E-mail: \{yyang\}@njust.edu.cn}
	\thanks{Yang Yang is with PCA Lab, Key Lab of Intelligent Perception and Systems for High-Dimensional Information of Ministry of Education, and Jiangsu Key Lab of Image and Video Understanding for Social Security, School of Computer Science and Engineering, Nanjing University of Science and Technology. Yang Yang is the corresponding author. 
	Yang Yang is also with Key Laboratory of Computer Network and Information Integration ( Southeast University ), Ministry of Education. State Key Lab. for Novel Software Technology, Nanjing University, P.R. China.}}

%
%

\markboth{Journal of \LaTeX\ Class Files,~Vol.~14, No.~8, August~2015}%
{Shell \MakeLowercase{\textit{et al.}}: Bare Demo of IEEEtran.cls for Computer Society Journals}
%



\IEEEtitleabstractindextext{%
\begin{abstract}
Image captioning can automatically generate captions for the given images, and the key challenge is to learn a mapping function from visual features to natural language features. Existing approaches are mostly supervised ones, i.e., each image has a corresponding sentence in the training set. However, considering that describing images always requires a huge of manpower, we usually have limited amount of described images (i.e., image-text pairs) and a large number of undescribed images in real-world applications. Thereby, a dilemma is the \emph{``Semi-Supervised Image Captioning''}. To solve this problem, we propose a novel Semi-Supervised Image Captioning method considering Wasserstein Graph Matching (SSIC-WGM), which turns to adopt the raw image inputs to supervise the generated sentences. Different from traditional single modal semi-supervised methods, the difficulty of semi-supervised cross-modal learning lies in constructing intermediately comparable information among heterogeneous modalities. In this paper, SSIC-WGM adopts the successful scene graphs as intermediate information, and constrains the generated sentences from two aspects: 1) inter-modal consistency. SSIC-WGM constructs the scene graphs of the raw image and generated sentence respectively, then employs the wasserstein distance to better measure the similarity between region embeddings of different graphs. 2) intra-modal consistency. SSIC-WGM takes the data augmentation techniques for the raw images, then constrains the consistency among augmented images and generated sentences. Consequently, SSIC-WGM combines the cross-modal pseudo supervision and structure invariant measure for efficiently using the undescribed images, and learns more reasonable mapping function. Experiments show that our method can outperform state-of-the-art comparison methods on the MS-COCO ``Karpathy'' offline test split, under different complex semi-supervised scenarios. 
\end{abstract}

}

\maketitle

\IEEEdisplaynontitleabstractindextext

%
\IEEEpeerreviewmaketitle

\IEEEraisesectionheading{\section{Introduction}\label{sec:introduction}}
\IEEEPARstart{I}{mage} captioning aims to automatically generate natural descriptions for the given images, which not only requires an insightful understanding of the input image's semantic contents, but also needs to express the information in a meaningful sentence~\cite{Hossain2018,KarpathyL15,ZhangP20}. Considering that natural language always plays an important role in many human activities but many applications only have visual input, thereby enabling machines to convert visual information into language has practical significance. For example, broadcasting road conditions by learning visual images to assist driving, producing natural human robot interactions, describing the product for e-commerce platform, and so on. Therefore, image captioning attracts increasing attentions and is becoming prominent in both academia and industry. In fact, the challenge of image captioning is to learn the mapping function between two heterogeneous modalities (i.e., visual features and linguistic features). In other words, model needs to recognize salient objects in an image using computer vision techniques, and integrate natural language processing to generate coherent descriptions correctly.

To this end, various approaches have been proposed. Early approaches mainly followed two lines, i.e., retrieval based and template based approaches~\cite{GuptaVJ12,OrdonezKB11,YangTDA11}. These methods accomplished the sentence generation by retrieving existing captions in the training set or relying on hand-coded language structures. However, the expressiveness of generated descriptions using these approaches is limited considering the inflexibility. Thanks to the advances of deep neural networks that are widely applied to the fields of computer vision~\cite{FarabetCNL13,IjjinaM16} and natural language processing~\cite{CollobertW08,BahdanauCB14,VaswaniSPUJGKP17}, deep neural network based image captioning methods are proposed and have demonstrated state-of-the-art results~\cite{MaoXYWY14a,XuBKCCSZB15,SammaniE19}. Inspired by recent advances in neural machine translation~\cite{ChoMGBBSB14}, researchers firstly explored the neural encoder-decoder models~\cite{KarpathyL15,YangYWCS16,yang2022exploiting}, which are composed of a CNN encoder and a LSTM decoder. Specifically, a CNN based encoder neural network first encodes an image into a set of intermediate representations, and then decodes the intermediate representations to a sentence word by word via a LSTM-based network. Furthermore, considering that we always need to mention the most salient contents in the description, approaches that utilizing attention mechanism~\cite{XuBKCCSZB15,LuXPS17,HuangWCW19} to guide description generation were proposed. These methods incorporated various kinds of attention mechanisms (for example, single or hierarchical attention) into the encoder–decoder framework, which enables the decoding process to focus on particular image region at each time step. However, \cite{SammaniM20} observed that prior image captioning models~\cite{XuBKCCSZB15,Anderson2018,YaoPLM18} learn global sentence/caption structure exceptionally well, but they often produce incorrect, inconsistent, or repetetive content. To mitigate this problem, several researches considered to edit inputs separately from the problem of generating inputs~\cite{HashimotoGOL18,SammaniE19}. Note that most current methods are assumed with supervised images in advance, i.e., all the images are with corresponding descriptions, which will take a huge of manpower. Nevertheless, it is relatively easy to access a large number of undescribed images in real-world applications. Thereby, a realistic challenge is the ``\textit{Semi-Supervised Image Captioning}'', which aims to conduct the captioning task in a semi-supervised scenario (i.e., limited described images and a huge number of undescribed images). 

To this end, unsupervised and semi-supervised image captioning approaches were proposed for handling undescribed images. Cognitive evidence~\cite{1983Vision} validates that visually grounded sentence generation largely attributed to the ``high-level'' symbolic reasoning in the semantic space. Based on this idea, \cite{Feng00L19a,GuJCZYW19} proposed unsupervised image captioning methods, which convert the images into text semantic space, and then use a discriminator with adversarial learning to evaluate the quality of generated sentences. However, the discriminator does not fundamentally evaluate the matching degree between the generated sentences and the images, so the performance is unsatisfactory. On the other hand, \cite{MithunPPR18,HuangKLCH19} developed semi-supervised style methods by directly utilizing undescribed images and their machine-generated sentences as the pseudo image-sentence pairs, together with described images to train the model. However, generating pseudo descriptions is more difficult comparing with predicting pseudo labels in semi-supervised classification task. In other words, the generator trained with limited described images will bring many noisy pseudo descriptions, which have negative effects on the training of mapping function in return. 

To sum up, the key challenge of semi-supervised image captioning is to \emph{design reasonable supervisions for qualifying the generated sentences}.  Actually, the input image and output sentence have similar semantic and structural information in the heterogeneous representation space for image captioning task. Therefore, the image input can be used as pseudo supervision to constrain the generated sentence in result, and the challenge of \emph{``reasonable supervision information''} can be transformed into finding cross-modal comparably representations. In fact, both the image and corresponding description can be abstracted into scene graphs with similar semantic information, thereby disentangling from the heterogeneous gap. Based on this thought, in this paper, we propose a novel Semi-Supervised Image Captioning method considering Wasserstein Graph Matching (SSIC-WGM), which incorporates the scene graphs as the intermediate representation into the encoder-decoder framework for semi-supervision, benefiting from the complementary strengths of both inter-modal and intra-modal consistencies. For inter-modal consistency, SSIC-WGM utilizes scene graphs~\cite{JohnsonGF18,WangLZY18}, which can be regarded as a unified representation that connects the objects, attributes, and relationships by directed edges, to bridge the gap between input image and generated sentence. To effectively measure the similarity of two scene graphs, SSIC-WGM combines the node embedding with the Wasserstein distance~\cite{Ludger2009Optimal,KolouriPTSR17} to measure the cross-modal consistency between two graphs inspired by~\cite{TogninalliGLRB19}. On the other hand, the sentences generated by the augmented images should also be similar, thereby SSIC-WGM employs data augmentation techniques~\cite{FrenchMF18} for conducting intra-modal consistency, which constrains the  consistency among generated sentences and augmented images. In result, we can effectively utilize the undescribed images and described ones in a unified semi-supervised framework. Note that SSIC-WGM can be implemented with any captioning model, and we adopt several typical approaches for verification~\cite{RennieMMRG17,ZhouWLHZ20}. Extensive experiments on MS-COCO~\cite{LinMBHPRDZ14} and FLICKR30K~\cite{YoungLHH14} validate the superiority of using SSIC-WGM in image captioning. In summary, our contributions in this paper are:
\begin{itemize}
	\item Propose a novel semi-supervised image captioning framework for handling undescribed images, which is universal for any captioning model;
	\item Propose the inter-modal consistency, which utilizes the symbolic scene graphs as the intermediate representations for heterogeneous image input and generated sentence, and employs the wasserstein distance to measure the similarity between two graphs;
	\item Propose the intra-modal consistency, which further constrains the consistency among generated sentences and augmented images.
\end{itemize}


\section{Related Work}
In this paper, we aim to develop the inter-modal consistency and intra-modal consistency for handling the undescribed images in image captioning task. Therefore, our work is related to: image captioning, scene graphs, and semi-supervised learning.

\subsection{Image Captioning}
Automatic image captioning has attracted wide attention in recent. Early works designed rules/templates based approaches~\cite{KuznetsovaOBBC12,LiKBBC11}, which generated slotted captioning templates manually, and then utilized the detected keywords to fill the templates~\cite{YaoYLLZ10}, but their expressive power was limited because of the need for designing templates manually. With the development of deep learning, modern captioning approaches have achieved breaking advances inspired from several NLP techniques, i.e., encoder-decoder technique~\cite{ChoMGBBSB14}, attention technique~\cite{HuangWCW19}. Encoder-decoder based methods are inspired by the neural machine translation. For example, \cite{MaoXYWY14a} directly modeled the probability distribution for generating a word given previous words and an image; \cite{VinyalsTBE15} proposed an end-to-end framework with a CNN encoding the image to feature vector and a LSTM decoding to caption; \cite{KarpathyL15} described a multi-modal recurrent neural network architecture that uses the inferred alignments to generate novel descriptions of image regions. Encouraged by the success of attention mechanism~\cite{BahdanauCB14}, \cite{XuBKCCSZB15} automatically learned to fix its gaze on salient objects while generating the corresponding words in the output sequence; \cite{YouJWFL16} learned a top-down and bottom-up computation selection, which selectively attended to semantic concept proposals and fused them into recurrent neural networks; \cite{HuangWCW19} added an attention-on-attention module after both the LSTM and the attention mechanism, which can measure the relevance between attention result and query. Besides, \cite{RennieMMRG17} considered the problem of optimizing image captioning systems using reinforcement learning, and showed the significant gains in performance. However, note that all these methods need a huge amount of described images for training, whereas the scenario with a large amount of undescribed images is more general in real applications. Thereby, \cite{MithunPPR18,HuangKLCH19} proposed to extract regional semantics from undescribed images as additional weak supervision. Nevertheless, the generated pseudo sentences are always unqualified to fine-tune the generator in real experiments. On the other hand, several attempts tried to employ the adversarial learning for unsupervised captioning, for example,  \cite{Feng00L19a} distilled the knowledge in visual concept detector into the captioning model to recognize the visual concepts, and adopted sentence corpus to teach the captioning model; \cite{GuJCZYW19} developed an unsupervised feature alignment method with adversarial learning that maps the scene graph features from the image to sentence modality. Nevertheless, these methods mainly depend on employing the domain discriminator for learning plausible sentences, which are difficult for generating matched sentences.

\subsection{Scene Graphs}
A scene graph is a unified representation that connects the objects (or entities), their attributes, and their pairwise relationships. It contains the structured semantic information and can be used for both image~\cite{JohnsonGF18} and text~\cite{WangLZY18}. Therefore, both the image and text can be transformed into the similar scene graphs, which provide beneficial prior for measuring the similarity between image and text modalities in return. At present, scene graph has been applied in various vision tasks like VQA~\cite{TeneyLH17}, image generation~\cite{JohnsonGF18}, and visual grounding~\cite{LiuZZW19}. The key challenge in semi-supervised image captioning is to effectively calculate the distance of two graphs. Thanks to the prominent success of graph neural networks (GNNs)~\cite{KipfW17} for graph-based data to learn more discriminative representations, we can use the GNN output, i.e., the feature vector that aggregates all nodes' embeddings using a readout function (for example, average pooling) as the final graph embedding, for calculating the similarity between graphs. Recently, wasserstein distance has been used for similarity learning on graphs~\cite{TogninalliGLRB19,Becigneul2020}, which calculates the similarity considering fine-grained region-wise embedding. For example, \cite{TogninalliGLRB19} proposed a wasserstein kernel for graphs that involves pairwise calculation of the wasserstein distance between graph representations by capturing the geometric information.

\subsection{Semi-Supervised Learning}
Traditional single modal semi-supervised learning always focuses on classification or clustering tasks, and mainly consider two aspects: 1) Entropy minimization~\cite{GrandvaletB04}, in which the model predictions of unlabeled data were converted to hard labels for calculating the cross-entropy~\cite{ArazoOAOM20,xi2023robust}. 2) Consistency regularization~\cite{BachmanAP14}, which produced random perturbations with data augmentation~\cite{FrenchMF18}, then enforced consistency between the augmentations~\cite{yang2021s2osc}. For example, \cite{XieDHL020} proposed unsupervised data augmentation with distribution alignment and augmentation anchoring, which encourages each output to be close to the weakly-augmented version of the same input; \cite{BerthelotCCKSZR20} used a weakly-augmented example to generate an artificial label and enforce consistency against strongly-augmented example. Similarly, existing multi-modal semi-supervised approaches also refer to these aspects, for example, Co-training~\cite{Ghani02} selected the most confident unlabeled instances to other modality for re-training; Co-regularization~\cite{sindhwani2008rkhs,yang2019comprehensive} constrained the consistency between the two modal predictions for each unlabeled instance. However, we always have no instance label information for image-text pairs, and using entropy minimization directly on generated sentence is more difficult than label prediction. Thereby, we turn to look for a more comfortable pseudo label (i.e., the scene graph) to supervise the generated sentences of undescribed images. 

The remainder of this paper is organized as follows. Section \ref{sec:s1} presents the proposed method. Section \ref{sec:s2} gives the experimental results on MS-COCO and FLICKR30K datasets, under different semi-supervised settings. Section \ref{sec:s3} concludes this paper.

\section{Semi-Supervised Image Captioning}\label{sec:s1}
In this section, we first give the definition of semi-supervised image captioning task and overall illustration. Then, we will provide the details of the proposed method, i.e., SSIC-WGM, which aims to effectively utilize the undescribed images.


\begin{figure*}[t]\centering
	\centering
	\includegraphics[width = 170mm]{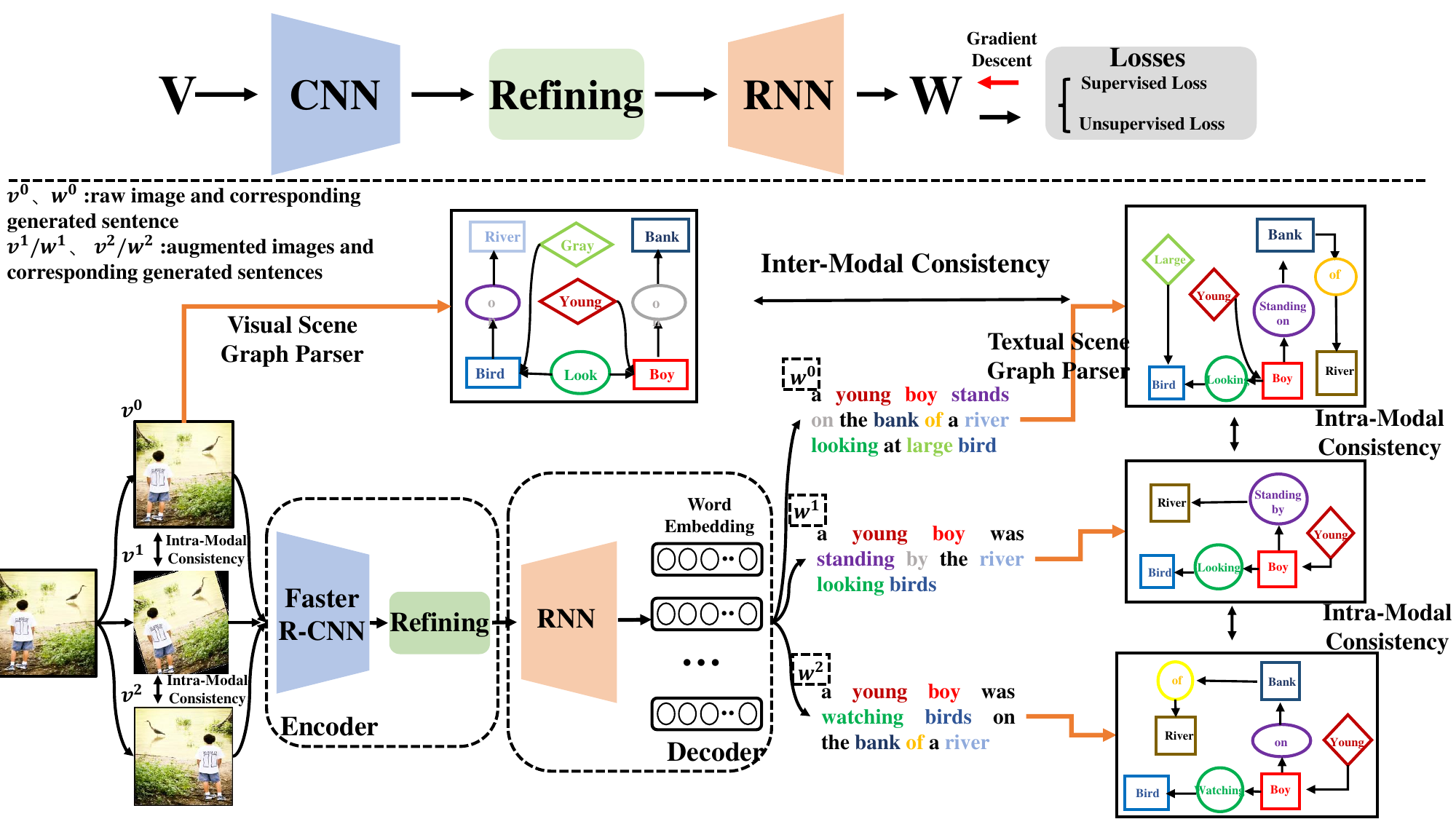}\\
	\caption{Diagram of the proposed SSIC-WGM. Top: the encoder-decoder model with semi-supervised loss. Bottom: our unsupervised loss, which qualify the generated sentences of undescribed images with intra-modal and inter-modal consistencies. In detail, two weakly-augmented images (i.e., $\vvv_1,\vvv_2$) and the raw image (i.e., $\vvv_0$) are fed into the encoder-decoder model to obtain corresponding sentences (i.e., $\w_0,\w_1,\w_2$). Inter-modal consistency calculates the distance between raw image's scene graph and generated sentence's scene graph, while intra-modal consistency constrains the distances between either two generated sentences'/images' scene graphs.}\label{fig:framework}
\end{figure*}

\subsection{Preliminaries}
In semi-supervised image captioning scenario, without any loss of generality, we define the image-sentence set as: $D = \{\{\vvv_i,\w_i\}_{i=1}^{N_l}, \{\vvv_j\}_{j=N_l + 1}^{N_l + N_u}\}$, where $\vvv_i \in \R^{d_v}$ denotes the $i-$th image, $\w_i \in \R^{d_w}$ represents the corresponding sentence. $N_l$ and $N_u$ ($N_l \ll N_u$) are the number of described and undescribed images, respectively. Different from traditional single modal tasks (for example, classification or clustering), image captioning is to learn a mapping function for automatically generating sentences using the input images, thereby the $\w_i$ acts as the supervision in the learning process. Following, we give the definition of semi-supervised image captioning:

\begin{defn}\label{def:d1}
	\textbf{Semi-Supervised Image Captioning} \emph{Given limited described images $\{\vvv_i,\w_i\}_{i=1}^{N_l}$ and a huge number of undescribed images $\{\vvv_j\}_{j=N_l + 1}^{N_l + N_u}$, we aim to construct a mapping function $G$ by effectively utilizing the undescribed images.}
\end{defn}

\subsection{Encoder-Decoder Model}
The key objective of image captioning is to learn the image-sentence generator, i.e., $G: \vvv \rightarrow \w $, which can effectively translate image to the sentence domain. In detail, given an image $\vvv$, the target of $G$ is to generate a natural language sentence $\hat{\w} = \{w_{1}, w_{2}, \cdots , w_{T} \}$ describing the image, $T$ denotes the length. Most of state-of-the-art captioning approaches~\cite{MaoXYWY14a,VinyalsTBE15,YouJWFL16,HuangWCW19} are based on encoder-decoder framework, which can be formulated as:
\begin{equation}\label{eq:e1}
\begin{split}
\hat{\w} = Decoder(Refining(Encoder(\vvv))), \\
\end{split}
\end{equation} 
where the encoder is usually a CNN~\cite{HeZRS16,RenHG017} based model for extracting the embedding of raw image input. The refining module is always attention mechanism~\cite{BahdanauCB14}, which aims to refine the visual embedding for suiting the language generation dynamically. The decoder is the widely used RNN-based model for sequence prediction $\hat{\w}$. Finally, the encoder, refining module and decoder compose the generator $G$. To train the $G$, we can minimize the cross-entropy loss (i.e., $\ell_{XE}$) or maximize the reinforcement learning based reward~\cite{RennieMMRG17} (i.e., $\ell_{RL}$), according to ground truth caption $\w$ and prediction $\hat{\w}$: 
\begin{equation}\label{eq:e2}
\begin{split}
\ell_{XE}(\theta)& = - \sum_{t = 1}^T \log p_{\theta}(\w_{t} | \w_{{1:t-1}}), \\
\ell_{RL}(\theta)&= - \EE_{\w_{1:T}}~ p_{\theta} [r(\w_{1:T})], \\
\end{split}
\end{equation} 
where $\w_{1:T}$ denotes the target ground truth sequence, the reward $r(\cdot)$ is a sentence-level metric for the generated sentence and the ground-truth, which always uses the score of captioning metric (e.g. CIDEr-D~\cite{VedantamZP15}). More details can refer to the supplementary. Without any loss of generality, the $G$ can be represented as any state-of-the-art model in our framework. In the experiments, we will combine our SSIC-WGM framework with various categories of captioning models to verify the effectiveness. 

Eq. \ref{eq:e2} reveals that all existing encoder-decoder models are supervised ones, i.e., all the images are with described sentences. However, there are a large number of undescribed images, i.e., $\{\vvv_j\}_{j=N_l + 1}^{N_l + N_u}$, in the semi-supervised scenario. Different from traditional single modal semi-supervised learning, the challenge of cross-modal semi-supervised learning is that the sequential predictions (i.e., generated sentences) are much more uncertain than multi-class predictions. In result, directly using the pseudo labeling as previous semi-supervised classification may lead negative effects for training.


\subsection{The Framework}
In the training process using semi-supervised image-sentence pairs, we first sample a mini-batch of images from the set $D$, including described and undescribed images. For described images, we can employ the loss function as Eq. \ref{eq:e2}. As to the undescribed images, we have no ground-truth for supervision. To solve this problem, we turn to utilize the raw image input for measuring the quality of generated sentences. As shown in Figure \ref{fig:framework}, on the one hand, inspired by~\cite{YangTZC19}, we extract the scene graphs as cross-modal intermediate representations for both the raw image and the generated sentence. On the other hand, we adopt the data augmentation techniques for each image (i.e., each image has $K$ variants), then acquire the corresponding generated sentences. The model is trained to maximize the consistency between inter-modal scene graphs, and the consistency among intra-modal of augmented images' and generated sentences' scene graphs.

Therefore, the proposed SSIC-WGM contains two main objects: 1) \textit{supervised loss}, which adopts the ground-truth sentences as the supervision for described images. 2) \textit{unsupervised loss}, which includes inter-modal and intra-modal consistency loss for undescribed images. The details of unsupervised loss are described below. 


\subsection{Inter-Modal Consistency}
The inter-modal consistency aims to calculate the distance between raw image input and generated sentence. A direct way is to calculate the distance of cross-modal feature representations, however, the gap between heterogeneous modalities is difficult to be effectively expressed by the similarity with the global embedding only. Thereby, inspired by~\cite{JohnsonGF18,WangLZY18}, SSIC-WGM turns to adopt a more efficient intermediate, i.e., scene graph. In detail, SSIC-WGM constructs the scene graphs for both the image and sentence using correspond techniques~\cite{JohnsonGF18,WangLZY18}, which can potentially transform the visually grounded representation and textually grounded representation into the shared semantic representation, so that we can shrink the heterogeneous gap and better measure the inter-modal distance. In the following parts, we will detail every component.

\subsubsection{Scene Graphs}
Formally, a scene graph consists a tuple $\G = \{\mathcal{V}, \E\}$~\cite{JohnsonGF18,WangLZY18}, where $\mathcal{V}$ and $\E$ are the sets of nodes and edges, respectively. $\mathcal{V}$ can be divided into three categories: object node $o$, attribute node $a$, and relationship node $r$. In detail, $o_m$ denotes the $m-$th object, $r_{mn}$ represents the relationship between $o_m$ and $o_n$, and $a_m$ is the attributes of $o_m$. Heterogeneous types of nodes are mapped into the common space, in other words, $o$, $a$ and $r$ are represented by $d-$dimensional vectors, i.e., ${\bf e}_o$, ${\bf e}_a$ and ${\bf e}_r$. For edges in $\E$, $a_m$ directly links to $o_m$ if object $o_m$ owns attributes $a_m$, $o_m$ links to $r_{mn}$ and $r_{mn}$ links to $o_n$ if $o_m$ and $o_n$ have relationship $r_{mn}$. We adopt the scene graph parser provided by~\cite{JohnsonGF18} for image modality, and parser provided by~\cite{WangLZY18} for sentence modality. Specifically, these parsers usually build a syntactic dependency tree from images or sentences, and then apply a rule-based method for transforming the tree to a scene graph. Consequently, for each image-sentence pair, we acquire the corresponding scene graphs $\G^v = P_v(\vvv)$ and $\G^w = P_w(\w)$, where $P_v$ and $P_w$ are the image parser and sentence parser, respectively.

\subsubsection{Wasserstein Distance on Graphs}
With the scene graphs $\G^v$ and $\G^w$, we aim to develop a distance function, i.e., $D(\G^v,\G^w)$, to measure the similarity between two graphs for supervising the parameter learning of $G$. A direct way can refer to the graph kernels~\cite{VishwanathanSKB10}, which rely on the $\R$-Convolution framework~\cite{Haussler1999}. These methods mainly adopt a naive aggregation (for example, average or sum) for all node embedding (for example, learned by the modern GCN~\cite{KipfW17}), so as to compare their substructures by calculating the Euclidean distance. However, this kind of methods has two drawbacks: 1) ignore node heterogeneity in scene graph. 2) mask important substructure differences between graphs. To overcome these problems, we first design a heterogeneous aggregation mechanism to conduct the node embedding, then calculate the wasserstein distance between the node embedding of two graphs, which aims to find subtler difference by considering graphs as high-dimensional objective rather than simple aggregations. The process has two steps: (1) transform each graph into a set of node embedding. (2) measure the distance between each pair of graphs. We first define an embedding scheme, then illustrate how to integrate embedding in the wasserstein distance.

\begin{defn}\label{def:d2}
	\textbf{Node Embedding.} \emph{Given a graph $\G = \{\mathcal{V}, \E\}$, the node embedding can be denoted as $f(\G) = \X$, where $f: \G \rightarrow \R^{|\mathcal{V}| \times d}$ is a function that outputs a $d-$dimensional representation for each node in the graph. The $i-$th row of $\X$ is the node embedding for $i-$th node in $\mathcal{V}$.}
\end{defn}

Traditional GNN~\cite{KipfW17} can be understood as special case of a simple differentiable message-passing framework, which is effective at accumulating and encoding features from local, structured neighborhoods. GNNs mainly concentrate on the aggregation of homogeneous nodes. However, the scene graph is a heterogeneous graph (i.e., with different types of nodes). Formally, $\X^v$ and $\X^w$ contains three kinds of $d$-dimensional embedding: object embedding $\x_{o}$ for object node $o$, attribute embedding $\x_{a}$ for attribute node $a$ and relationship embedding $\x_{r}$ for relationship node $r$. Thereby, we develop a node-specific transformation for $f$, which calculates the forward-pass update of a node considering the type of a neighbors:
\begin{equation}\label{eq:e3}
\begin{split}
h_m^{l+1} = \sigma(\sum_{k \in N_m^{nt}} W^l_{nt}h^l_k + h_m^{l})
\end{split}
\end{equation} 
where $N_m^{nt}$ denotes the set of neighbor indices of node $m$ under node type $nt \in \{o, r, a\}$. In detail, consider the linkage and physical meaning of neighbors as~\cite{YangTZC19}, the $nt$ of $r$ is $o$, the $nt$ of $a$ is $o$, the $nt$ of $o$ are $r$ and $a$. $h_m^{l}$ denotes the hidden representations of $m-$th node on $l-$th layer.  $W^l_{nt} \in \R^{d \times d}$ is the specific weight matrix for $nt$. $\sigma$ represents the element-wise activation function, for example, $ReLU(\cdot) = \max(0,\cdot)$. It is notable that  $\G^v$ and $\G^w$ can share the same transformation structure. Despite the above parametric embedding process in Eq. \ref{eq:e3}, we can also adopt the non-parametric form:
\begin{equation}\label{eq:e4}
\begin{split}
h_m^{l+1} = \sum_{k \in N_m^{nt}} \alpha_{k,m}^l h^l_k + h_m^{l}
\end{split}
\end{equation} 
which directly aggregates the neighbors without learnable parameters, $\alpha_{k,m}^l = \frac{d(h^l_k,h_m^{l})}{\sum_{j \in N_m^{nt}} d(h^l_j,h_m^{l})}$ is the weight for $h^l_k$ in weighted aggregation, $d$ measures the distance between two nodes, and we adopt squared Euclidean here. One of the appealing advantages is its simplicity, as there is no trainable parameter involved. We will compare these two forms in the experiments. In result, we can achieve the $\X^v \in \R^{|\mathcal{V}_v| \times d}$ and $\X^w \in \R^{|\mathcal{V}_w| \times d}$. Following, we will discuss how to measure the distance between graphs.

\begin{defn}\label{def:d3}
	\textbf{Graph Wasserstein Distance} \emph{Given two graph $\G$ and $\G'$, a node embedding function $f: \G \rightarrow \R^{|V| \times d}$, the Graph Wasserstein Distance (GWD) can be defined as $D_W^f(\G,\G') = W(f(\G),f(\G'))$}
\end{defn}

Here, different from traditional wasserstein distance considering continuous probability distributions, we turn to deal with finite sets of node embedding. Therefore, we can reformulate the wasserstein distance as a sum rather than an integral, and use the matrix notation commonly encountered in the optimal transport~\cite{RubnerTG00} to represent the transportation plan. Consequently, given two sets of vectors $\X^v$ and $\X^w$, we can define the wasserstein distance as:
\begin{equation}\label{eq:e5}
\begin{split}
W (\X^v, \X^w) = & \min\limits_{T_{inter} \in \varGamma (\X^v, \X^w)} <T_{inter},M_{inter}> ,\\
s.t. T_{inter} {\bf 1}_{|\mathcal{V}_v|} = & {\bf 1}_{|\mathcal{V}_w|}/|\mathcal{V}_w| \\
T_{inter}^\top {\bf 1}_{|\mathcal{V}_w|} = & {\bf 1}_{|\mathcal{V}_v|}/|\mathcal{V}_v| \\
\end{split}
\end{equation} 
where $M_{inter}$ is the distance matrix containing the distances $d(\x^v,\x^w)$ between each element $\x^v$ of $\X^v$ and $\x^w$ of $\X^w$, we utilize the squared Euclidean distance for $d$ here. $T_{inter} \in \varGamma$ is a transport matrix (or joint probability), which represents the node correspondences between $\G^v$ and $\G^w$. $<\cdot,\cdot>$ is the Frobenius dot product. The total mass to be transported is distributed across the elements of $\X^v$ and $\X^w$, thereby the row/column values of $T$ must sum up to ${\bf 1}_{|\mathcal{V}_w|}/|\mathcal{V}_w|$/${\bf 1}_{|\mathcal{V}_v|}/|\mathcal{V}_v|$, respectively.
In summary, the transport matrix $T$ contains the fractions that indicates how to transport the values from $\X^v$ to $\X^w$ with the minimal total transport effort, in other words, $W (\X^v, \X^w)$ can reflect the distance between two scene graphs. 

\subsection{Intra-Modal Consistency}
In addition to using inter-modal consistency for generated sentences, we can refer to the intra-modal consistency to supervise the generated sentences. In detail, we create several variants for each input image with data augmentation techniques, then constrain the consistency among the corresponding generated sentences.

Traditional consistency training in semi-supervised learning always regularizes model predictions invariant to augmented examples. This framework is useful because a good model should be robust to any small change. Existing methods usually leverage two kinds of augmentations: 1) Weak augmentation is a standard flip-and-shift strategy, which does not significantly change the content of the input. 2) Strong augmentation always refers to the AutoAugment~\cite{Cubuk20} and its variants, which use reinforcement learning to find an augmentation
strategy comprising transformations from the Python Imaging Library\footnote{https://www.pythonware.com/products/pil/}. Therefore, each image input can be denoted as a bag of $K+1$ instances, i.e., $\Psi(\vvv_j)$, and the corresponding generated sentences can also be represented as a bag of instances, i.e., $G(\Psi(\vvv_j))$. 


Different from the traditional semi-supervised methods that regularize the consistency of label prediction, the intra-modal consistency in image captioning task has two challenges: 1) Fine-grained consistency. The task of image captioning is to generate a sequential words, rather than simple prediction. However, the strong augmentation that may cut out image will impact the quality of generated sentences. 2) Sequential consistency. Image captioning task needs to constrain the consistency among generated sentences. Actually, it is unsuitable to directly regularize the sequential feature or prediction consistency between any two sentences $\hat{\w}, \hat{\w}' \in G(\Psi(\vvv_j))$, because the words' sequence in $\hat{\w}, \hat{\w}'$ may be different, which may cause extra noises. To overcome these problems, we select the weak augmentation, and adopt the scene graph comparison as inter-modal consistency, which can comprehensively consider the position and embedding of words in sentence. In result, the intra-modal consistency between any two sentences can be represented as:
\begin{equation}\label{eq:e6}
\begin{split}
W (\X^w, \X^{w'}) = & \min\limits_{T_{tra} \in \varGamma (\X^w, \X^{w'})} <T_{tra},M_{tra}> ,\\
s.t. T_{tra} {\bf 1}_{|\mathcal{V}_{w}|} = & {\bf 1}_{|\mathcal{V}_{w'}|}/|\mathcal{V}_{w'}| \\
T_{tra}^\top {\bf 1}_{|\mathcal{V}_{w'}|} = & {\bf 1}_{|\mathcal{V}_{w}|}/|\mathcal{V}_{w}|  \\
\end{split}
\end{equation} 
where $M_{tra}$ is the distance matrix between $\X^w$ and $\X^{w'}$, we also utilize the squared Euclidean distance for calculation. $T_{tra} \in \varGamma$ is the transport matrix. The intra-modal consistency of images is the same.

\subsection{Overall Objective}
In summary, with the limited amount of described images and large amount of undescribed images, we define the total loss by combining the Eq. \ref{eq:e2}, Eq. \ref{eq:e5} and Eq. \ref{eq:e6}:
\begin{equation}\label{eq:e7}
\begin{split}
L = & L_c +  \lambda_1 L_{inter} + \lambda_2 L_{intra}, \\
L_{inter} = & \sum_{j=1}^{N_u} W (\X^v_j, \X^w_j), \\
L_{intra} = & \sum_{j=1}^{N_u} \sum_{m \neq n \in \{1: K+1\}} W (\X^w_{j_m}, \X^w_{j_n}),
\end{split}
\end{equation}
where $L_c$ denotes the captioning loss, which can be adopted as $\ell_{XE}$ or $\ell_{RL}$ in Eq. \ref{eq:e2}. $\lambda_1$ and $\lambda_2$ are scale values that control the weights of different losses. The constraint is the same as Eq. \ref{eq:e5} and Eq. \ref{eq:e6}. The first term is the original supervised loss, and the last two terms are unsupervised loss for utilizing the undescribed images. Furthermore, we can add the augmented images' intra-modal consistency to constitute the SSIC-WGM+, in which the $L_{intra}$ can be reformulated as:
\begin{equation}\label{eq:e9}
\begin{split}
L_{intra} = & \sum_{j=1}^{N_u} \sum_{m \neq n \in \{1: K+1\}} W (\X^v_{j_m}, \X^v_{j_n}) + W (\X^w_{j_m}, \X^w_{j_n}),
\end{split}
\end{equation}

The parameters in Eq. \ref{eq:e7} include: $T_{inter}$ , $T_{tra}^v$, $T_{tra}^w$ and $G$. Actually, the solution of transport matrix  $T_{inter}$ , $T_{tra}^v$ and $T_{tra}^w$ has closed-form in the forward process by using the $M_{inter}/M_{tra}^v/M_{tra}^w$. In detail, with calculated $M_{inter}/M_{tra}^v/M_{tra}^w$, $T_{inter}/T_{tra}^v/T_{tra}^w$ is the solution of an entropy-smoothed optimal transport problem ($T_{inter}/T_{tra}^v/T_{tra}^w$ has the same solution, so we utilize the $T$ as substitution):
\begin{equation}\label{eq:e8}
\begin{split}
T = \arg\min\limits_{T} \lambda \langle T, M \rangle  - \varOmega(T)
\end{split}
\end{equation}
here $\varOmega(T) = -\sum_{a,b} T_{a,b} log T_{a,b} $ is the entropy of $T$, and $\lambda > 0$ is entropic regularization coefficient. $T_{a,b}$ denotes the element in the $a-$th row and $b-$th column of $T$. Based on the Sinkhorn theorem, we conclude that the transportation matrix can be written in the form of $T^\star = diag(u) K diag(v)$, where $K = exp(-\lambda M)$ is the element-wise exponential of $\lambda M-1$. Besides, in Sinkhorn iterations, $u$ and $v$ are kept updating. Taking the k-th iteration as an example, the update takes the following forms:$u = \frac{{\bf 1}_{|\mathcal{V}_{w}|}/|\mathcal{V}_{w}|}{K^\top u^{k-1}}$ and $v = \frac{{\bf 1}_{|\mathcal{V}_{w'}|}/|\mathcal{V}_{w'}|}{K^\top v^{k-1}}$. After calculating the $T_{inter}$, $T_{tra}^v$ and $T_{tra}^w$, we can directly adopt the gradient descent technique to update parameters of $G$ considering the loss $L$. In training process, it is notable that cross-entropy and reinforcement learning optimization are applied for supervised data, and most supervised image captioning approaches~\cite{HuangKLCH19,XuNTLD021,ZhouWLHZ20,HerdadeKBS19,RennieMMRG17,HuangWXC19,HuangWCW19} use reinforcement learning optimization to further fine-tune the mode trained with cross-entropy loss using supervised data. In our framework, we directly utilize the semi-supervised loss in cross-entropy training phase. As for CIDEr-D Score Optimization setting, we firstly adopt the semi-supervised loss (including the cross-entropy loss for supervised data) for pre-training, then fine-tune by replacing the cross-entropy with reinforcement learning loss in the semi-supervised loss. Therefore, the proposed semi-supervised method can function at both the cross-entropy training and reinforcement learning stages.

\begin{table*}[!htb]{
		\centering
		\caption{Performance of comparison methods on MS-COCO “Karpathy” test split, where B$@$N, M, R, C and S are short for BLEU@N, METEOR, ROUGE-L, CIDEr-D and SPICE scores.}
		\label{tab:tab1}
		\begin{tabular*}{1\textwidth}{@{\extracolsep{\fill}}@{}l@{}|@{}c@{}|@{}c@{}|@{}c@{}|@{}c@{}|@{}c@{}|@{}c@{}|@{}c@{}|@{}c@{}|@{}c@{}|@{}c@{}|@{}c@{}|@{}c@{}|@{}c@{}|@{}c@{}|@{}c@{}|@{}c@{}}
			\toprule
			\multirow{2}{*}{Methods} & \multicolumn{8}{c|}{Cross-Entropy Loss} & \multicolumn{8}{c}{CIDEr-D Score Optimization} \\
			\cmidrule(l){2-17}
			& B$@$1 & B$@$2 & B$@$3 & B$@$4 & M & R & C & S & B$@$1 & B$@$2 & B$@$3 & B$@$4 & M & R & C & S\\
			\midrule
			SCST  &56.8&38.6&25.4&16.3&16.0 &42.4&38.9&9.3&59.4 &39.5&25.3 &16.3&17.0&42.9&43.7&9.9\\
			AoANet &67.9&49.8&34.7 &23.2&20.9&49.2&69.2&14.3&66.8 &48.6&34.1&23.6&21.8 &48.7&70.4&15.2\\
			AAT &63.2&45.8&31.7 &21.3&19.0&47.6&58.0&12.4&66.7 &48.1&33.3&22.7&20.4 &47.8&63.5&13.2\\
			ORT &63.6&45.8&31.7 &21.4&19.4&46.9&61.1&12.6&65.3 &46.5&31.9&21.3&20.3 &47.2&62.0&13.3\\
			GIC &63.0&46.8&33.2 &20.0&19.2&50.3&50.5&12.3&64.7 &46.9&32.0&20.7&19.0 &47.8&55.7&12.5\\
			Ancaption &56.5&38.1&24.9 &16.5&16.2&42.2&40.5&10.5&57.2 &38.9&25.5&17.3&16.7 &42.9&45.6&10.7\\
			RSTNet &68.1 &50.2 &34.8 &23.4 &20.9 &49.6 &69.5 &14.3 &67.3 &48.9 &34.5 &23.6 &22.0 &48.8 &70.7 &15.3\\
			\midrule
			Graph-align &-&-&-&-&-&-&-&-&67.1 &47.8&32.3&21.5&20.9&47.2&69.5&15.0\\
			UIC &-&-&-&-&-&-&-&-&41.0 &22.5&11.2&5.6&12.4&28.7&28.6&8.1\\
			R2M &-&-&-&-&-&-&-&-&44.0&25.4&12.7&6.4&13.0&31.3&29.0&9.1\\
			RSA &-&-&-&-&-&-&-&-&49.5&27.3&13.1&6.3&14.0&34.5&31.9&8.6\\
			SME &-&-&-&-&-&-&-&-&-&-&-&6.5&12.9&35.1&22.7&7.4\\
			SCS &-&-&-&-&-&-&-&-&67.1&47.9&33.4&22.8&21.4&47.7&74.7&15.1\\
			\midrule
			A3VSE &68.0&50.0&34.9 &23.3&20.8&49.3&69.6&14.5&67.6 &49.6&35.2&24.5&22.1 &49.3&72.4&15.3\\
			Self-Dis &-&-&-&-&-&-&-&- &67.9&49.8&35.4&25.0&21.7&49.3&73.0&14.5\\
			Per-Pre &-&-&-&-&-&-&-&-&55.0&37.8&25.8&17.8&16.1&41.5&41.1&8.3\\
			\midrule
			AoANet+Inter &68.3&50.4&35.8&24.7&22.6&49.7&72.8&15.0&68.3&50.7&36.3&25.1&22.9&50.8&73.9&15.9\\
			AoANet+Intra &68.1&50.1&35.3&24.5&22.5&49.4&72.3&14.9&68.1&50.2&35.9&24.8&22.7&50.4&73.6&15.8\\
			AoANet+  &68.4 &50.4 &35.9 &25.0 &22.8 &50.1 &73.0 &15.1 &68.6 &50.9 &36.4 &25.4 &23.0 &50.9 &74.2 &16.1\\
			SSL &68.1&50.2&35.1&23.8&21.6&49.4&71.9&14.8&68.2&50.2&35.9&24.8&22.5&50.2&72.7&15.4\\
			Strong+ &68.5&50.5&35.6&24.7&22.5&49.7&72.7&15.0&68.6&50.8&35.9&25.1&22.8&50.7&73.7&15.5\\
			GCN+  &68.0&50.1&34.9&23.5&20.9&49.3&69.7&14.5&68.1&50.3&35.3&24.4&22.3&49.5&72.6&15.2\\
			w/o Inter &68.5&50.7&35.8&24.8&22.8&50.3&73.1&15.3&68.5&50.9&36.1&25.1&23.0&50.6&74.0&16.1\\
			w/o Intra &68.6&50.9&35.9&25.0&23.0&50.5&73.7&15.6&68.9&51.4&36.4&25.5&23.1&51.2&75.2&16.3\\
			\midrule			
			SSIC-WGM &\bf68.8&51.3&36.1&25.3&23.1&50.9&74.1&15.7&69.1&51.7&\bf36.7&25.8&23.2&51.5&75.4&\bf16.5\\
			SSIC-WGM+  &68.8 &\bf 51.4 &\bf36.3 &\bf25.4 &\bf23.3 &\bf51.0 &\bf74.6 &\bf15.8 &\bf69.2 &\bf51.8 &36.7  &\bf26.1 &\bf23.5 &\bf51.8 &\bf76.0 &16.5\\
			\bottomrule
	\end{tabular*}}
\end{table*}

\begin{table*}[!htb]{
		\centering
		\caption{Performance of comparison methods on FLICKR30K dataset, where B$@$N, M, R, C and S are short for BLEU@N, METEOR, ROUGE-L, CIDEr-D and SPICE scores. }
		\label{tab:tab4}
		\begin{tabular*}{1\textwidth}{@{\extracolsep{\fill}}@{}l@{}|@{}c@{}|@{}c@{}|@{}c@{}|@{}c@{}|@{}c@{}|@{}c@{}|@{}c@{}|@{}c@{}|@{}c@{}|@{}c@{}|@{}c@{}|@{}c@{}|@{}c@{}|@{}c@{}|@{}c@{}|@{}c@{}}
			\toprule
			\multirow{2}{*}{Methods} & \multicolumn{8}{c|}{Cross-Entropy Loss} & \multicolumn{8}{c}{CIDEr-D Score Optimization} \\
			\cmidrule(l){2-17}
			& B$@$1 & B$@$2 & B$@$3 & B$@$4 & M & R & C & S & B$@$1 & B$@$2 & B$@$3 & B$@$4 & M & R & C & S\\
			\midrule
			SCST  &35.5 &21.0 &12.5  &7.7 &11.3 &31.7 &7.1  &7.1 &38.2 &22.9 &13.8  &8.6 &11.7 &32.8 &8.3  &7.4\\
			AoANet &55.2 &35.8 &22.7  &14.2 &15.7 &39.4 &24.5  &10.1 &58.9 &38.5 &24.3  &15.1 &15.0 &39.9 &23.9  &9.2\\
			AAT &53.9 &34.6 &21.0  &13.0 &15.0 &38.6 &19.5  &9.3 &52.5 &33.1 &19.7  &11.8 &14.0 &35.4 &18.5  &8.9\\
			ORT &54.3 &34.9 &21.5  &13.5 &15.2 &38.9 &23.1  &9.4 &56.9 &37.3 &22.5  &14.2 &14.8 &38.6 &22.4  &9.1\\
			GIC &34.7 &20.5 &12.0  &7.3 &10.8 &30.5 &7.0  &6.8 &37.6 &22.1 &13.6  &8.4 &11.4 &31.6 &8.3  &7.5\\
			Ancaption &35.2 &20.8 &12.1  &7.5 &11.0 &30.8 &7.1  &6.8 &38.0 &22.6 &13.6  &8.4 &11.2 &32.6 &8.1  &7.3\\
			RSTNet &55.6 &35.8 &22.9  &14.6 &15.8 &39.7 &24.8  &10.2 &55.4 &35.3 &22.5  &14.5 &15.6 &39.5 &24.2 &9.5\\
			\midrule
			Graph-align &-&-&-&-&-&-&-&- &-&-&-&-&-&-&-&-\\
			UIC &-&-&-&-&-&-&-&- &-&-&-&-&-&-&-&-\\
			R2M &-&-&-&-&-&-&-&-&53.1&32.8&19.2&11.7&13.7&35.9&18.1&8.3\\
			RSA &-&-&-&-&-&-&-&-&54.6&33.5&20.5&11.9&14.3&37.4&20.4&8.5\\
			SME &-&-&-&-&-&-&-&-&-&-&-&7.9&13.0&32.8&9.9&7.5\\
			SCS &-&-&-&-&-&-&-&-&-&-&-&14.3&15.6&38.5&20.5&-\\
			\midrule
			A3VSE &56.6 &37.1 &23.7  &15.0 &15.7 &39.7 &25.4  &10.1 &56.3 &36.8 &23.5  &14.8 &15.9 &39.6 &25.4  &10.2\\
			Self-Dis &-&-&-&-&-&-&-&- &53.0&32.7&20.5&13.4&13.7&36.0&13.1&8.0\\
			Per-Pre &-&-&-&-&-&-&-&- &48.0&29.0&16.5&10.1&11.6&33.0&10.0&5.8\\
			\midrule
			AoANet+Inter &57.0 &37.5 &23.9 &15.2 &15.8 &39.9 &26.1 &10.3 &57.9 &38.3 &24.5 &15.5 &16.0 &40.6 &26.5 &10.5\\
			AoANet+Intra &56.7 &37.4 &23.7 &15.0 &15.7 &39.8 &25.7 &10.2 &57.5 &38.1 &24.2 &15.4 &15.9 &40.2 &26.3 &10.3\\
			AoANet+  &57.0 &37.5 &24.0 &15.2 &15.8 &39.9 &26.3 &10.3 &58.1 &38.5 &24.6 &15.5 &16.2 &40.8 &26.8 &10.5\\
			SSL &56.8 &37.2 &23.7 &15.1 &15.7 &39.8 &25.6 &10.1 &57.2 &37.5 &23.9 &15.2 &15.9 &40.0 &25.9 &10.3\\
			Strong+ &57.2 &37.5 &24.0 &15.2 &15.8 &40.2 &26.4 &10.3 &58.4 &38.5 &25.1 &15.7 &16.3 &40.8 &26.7 &10.5\\
			GCN+  &56.7 &37.2 &23.8 &15.1 &15.7 &39.8 &25.5 &10.1 &57.0 &37.4 &23.9 &15.3 &15.9 &39.8 &25.7 &10.3\\
			w/o Inter &57.0 &37.5 &24.0 &15.2 &15.8 &40.1 &26.3 &10.3 &58.3 &38.5 &25.1 &15.6 &16.3 &40.8 &26.7 &10.5\\
			w/o Intra &57.2 &37.6 &24.2 &15.3 &15.8 &40.2 &26.5 &10.4 &58.6 &38.6 &25.3 &15.7 &16.3 &40.9 &26.9 &10.6\\
			\midrule
			SSIC-WGM & 57.3 &\bf 37.8 &\bf 24.3 & 15.4 & 15.9 &\bf 40.2 & 26.7 &\bf 10.5 & 58.9 & 38.7 & 25.6 &\bf 15.8 & 16.4 & 41.0 & 27.1 & 10.7\\
			SSIC-WGM+  &\bf57.4 &37.8 &24.3 &\bf15.5 &\bf16.0 &40.2 &\bf27.1 &10.5 &\bf59.1 &\bf38.9 &\bf25.7 &15.8 &\bf16.6 &\bf41.3 &\bf27.6 &\bf10.8\\
			\bottomrule
	\end{tabular*}}
\end{table*}


\section{Experiments}\label{sec:s2}
\subsection{Datasets}
Following~\cite{HuangWCW19,HuangWXC19,HerdadeKBS19,ZhouWLHZ20,RennieMMRG17}, we adopt the popular MS-COCO dataset~\cite{LinMBHPRDZ14} and FLICKR30K~\cite{YoungLHH14} for evaluation, which are common datasets for image captioning. In detail, MS-COCO contains 123,287 images (82,783 training images and 40,504 validation images), each labeled with 5 captions. According to the split of test set, we adopt the offline ``Karpathy'' data split~\cite{KarpathyF17} contains 5,000 images for validation, 5,000 images for testing, and the rests are for training. FLICKR30K consists 31,014 photographs of everyday activities, events and scenes (all harvested from Flickr) and 158,915 captions (obtained via crowd sourcing), the offline ``Karpathy'' data split~\cite{KarpathyF17} contains 29,000 training images, 1,014 validation images and 1,000 images for testing. To construct the semi-supervised scenario, we randomly select examples with artificially set proportions from the training set as supervised data, and the rests are unsupervised data.

\subsection{Baselines and Evaluation protocol}
We compare our model with the state-of-the-art captioning approaches.  The comparison models fall into three categories: 1) state-of-the-art supervised captioning methods: SCST~\cite{RennieMMRG17}, AoANet~\cite{HuangWCW19}, AAT~\cite{HuangWXC19}, ORT~\cite{HerdadeKBS19}, GIC~\cite{ZhouWLHZ20}, Ancaption~\cite{Xu2021}, and RSTNet~\cite{ZhangSLJZWHJ21}. It is notable that supervised captioning methods can only develop the mapping functions with supervised data, and leave out the unsupervised data. 2) state-of-the-art unsupervised captioning methods: Graph-align~\cite{GuJCZYW19} UIC~\cite{Feng00L19a}, R2M~\cite{guo2020recurrent}, RSA~\cite{honda-etal-2021-removing}, SME~\cite{laina2019towards} and SCS~\cite{ben2021unpaired}. These approaches utilize the independent image set and corpus set for training. 3) state-of-the-art semi-supervised method: A3VSE~\cite{HuangKLCH19}, Self-Dis~\cite{ChenJZ21} and Per-Pre~\cite{JainSJM021}. In detail, the A3VSE is designed for cross-model retrieval, we borrow its framework to generate pseudo sentences for second training the mapping function. 

For evaluation, we use different metrics, including BLEU~\cite{PapineniRWZ02}, METEOR~\cite{BanerjeeL05}, ROUGE-L~\cite{HuangWCW19}, CIDEr-D~\cite{VedantamZP15} and SPICE~\cite{AndersonFJG16}, to evaluate the proposed method and comparison methods. All the metrics are computed with the publicly released code\footnote{https://github.com/tylin/coco-caption}, where B$@$N, M, R, C and S are short for BLEU@N, METEOR, ROUGE-L, CIDEr-D and SPICE score. It is notable that the CIDEr-D and SPICE metric is more suitable for the image captioning task~\cite{AndersonFJG16,VedantamZP15}. One of the problems with BLEU, ROUGE-L, CIDEr-D and METEOR is that these metrics are sensitive to n-gram overlap, which is neither necessary nor sufficient for two sentences to convey the same meaning~\cite{GimenezM07a}.

\subsection{Quantitative Analysis}
We first display the captioning performance under semi-supervised setting. Table \ref{tab:tab1} presents the comparison results on MS-COCO dataset, under the scenario using 1$\%$ supervised and 99$\%$ unsupervised data for the training set. For fairness, all the models are first trained under cross-entropy loss and then optimized for CIDEr-D score following~\cite{HuangWCW19}. Considering the nature of task, we directly adopt the given results of unsupervised methods, while rerun or modify the supervised/semi-supervised methods to adapt to the semi-supervised scenario, and adjust the parameters for best performance according to the raw paper. ``-'' represents the results have not been given in the raw paper. The results reveal that: 1) UIC achieves the worst performance on all metrics. This verifies that the generated sentence may mismatch the image with a high probability when only considering the domain discriminator. Graph-align and SCS methods perform better, as it introduces the scene graph or pre-defined dictionary into the discriminator for learning mapping function, which can better minimize the heterogeneous gap and well consider fine-grained matching. The difference between Graph-align and our method is that Graph-align still uses the idea of discriminator to measure the distribution difference, rather than to measure the specific example matching. Thereby, Graph-align performs worse than A3VSE and SSIC-WGM on most metrics. 2) A3VSE and Self-Dis show effects on improving the captioning performance, for example, cross-entropy loss/CIDEr-D score optimization of A3VSE improves 0.4/2.0 and 0.2/0.1 on CIDEr-D and SPICE scores when comparing with AoANet, but the improvement is limited because it is more difficult to ensure the quality of generated sentences. Per-Pre is a transductive method, i.e., the test set is also used during model training, so there is a decrease in performance when inductive training. 3) SSIC-WGM and SSIC-WGM+ achieve the highest scores among all compared methods in terms of all metrics, on both the cross-entropy loss and CIDEr-D score optimization. For example, SSIC-WGM+ achieves a state-of-the-art performance of 74.6/76.0 (CIDEr-D score) and 15.8/16.5 (SPICE score) under two losses (cross-entropy and CIDEr-D score), that acquire 5.4/5.6 and 1.5/1.3 improvements comparing with AoANet. The phenomena indicates that, with limited amount of supervised data, existing methods cannot construct a well mapping function, whereas SSIC-WGM and SSIC-WGM+ can reliably utilize the undescribed image to enhance the model. Table \ref{tab:tab4} records the results of FLICKR30K dataset, we can obtain conclusions similar to the MS-COCO dataset, thus verifying the effectiveness of SSIC-WGM and SSIC-WGM+ in different datasets. 

\begin{figure}[!htb]
	\begin{center}
		\begin{minipage}[h]{44mm}
			\centering
			\includegraphics[width=44mm]{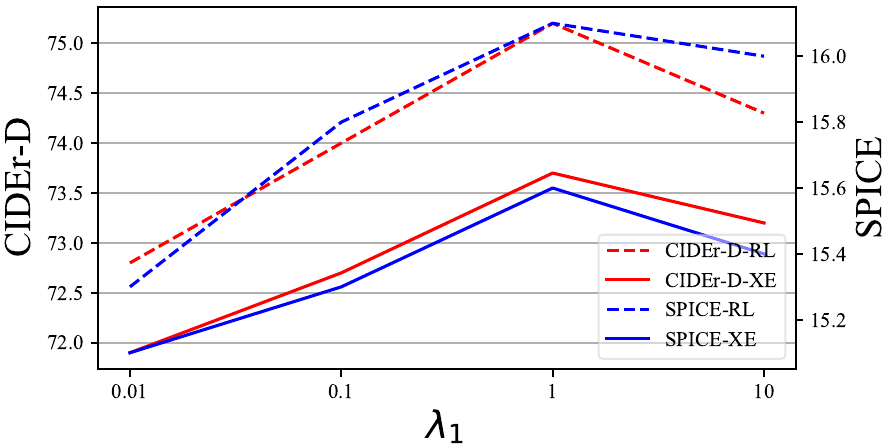}\\
			\mbox{ \;\;\;\; ({\it a}) {\small Inter-Modal Consistency}}
		\end{minipage}
		\begin{minipage}[h]{44mm}
			\centering
			\includegraphics[width=44mm]{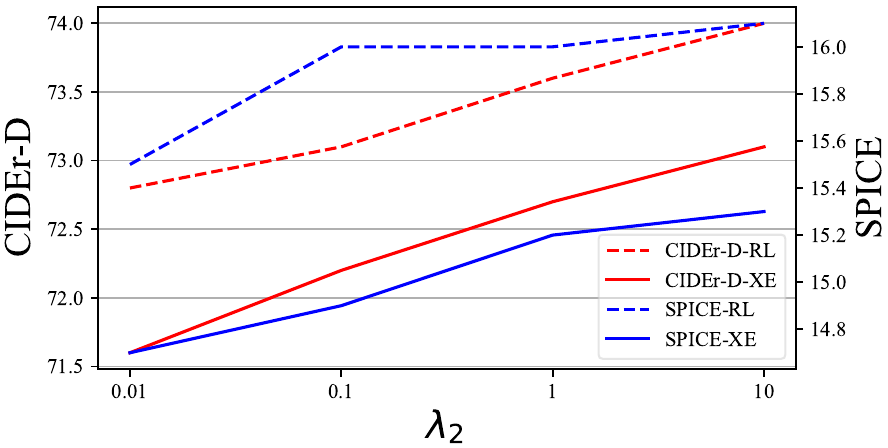}\\
			\mbox{ \;\;\;\; ({\it b}) {\small Intra-Modal Consistency}}
		\end{minipage}
	\end{center}
	\caption{Effect analyses of Inter-Modal Consistency and Intra-Modal Consistency with Cross-Entropy Loss.}\label{fig:f10}
\end{figure}

\begin{figure*}[!htb]
	\begin{center}
		\begin{minipage}[h]{43mm}
			\centering
			\includegraphics[width=43mm]{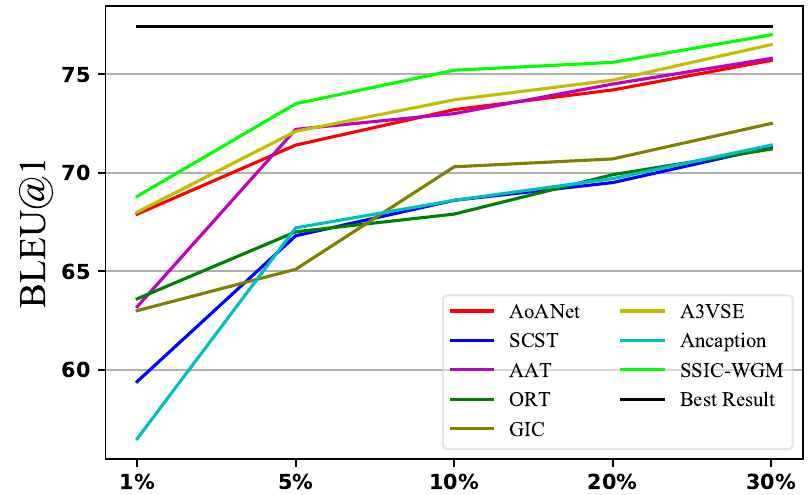}\\
			\mbox{ \;\;\;\; ({\it a1}) {BLEU$@$1 (XE)}}
		\end{minipage}
		\begin{minipage}[h]{43mm}
			\centering
			\includegraphics[width=43mm]{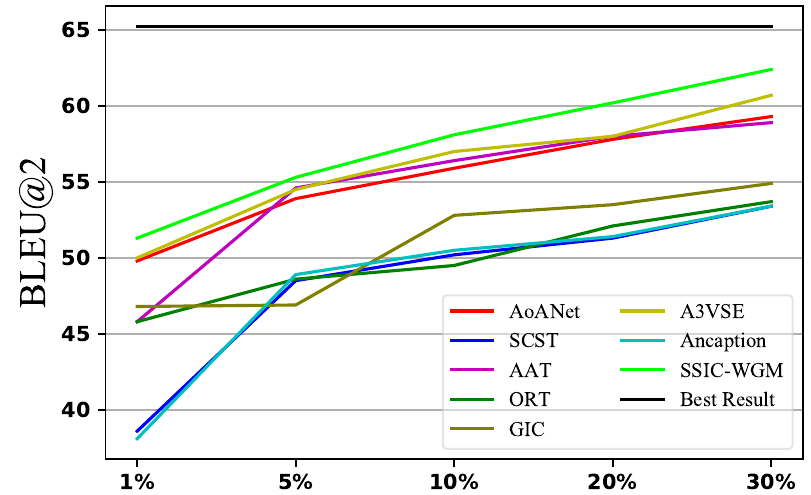}\\
			\mbox{ \;\;\;\; ({\it b1}) {BLEU$@$2 (XE)}}
		\end{minipage} 
		\begin{minipage}[h]{43mm}
			\centering
			\includegraphics[width=43mm]{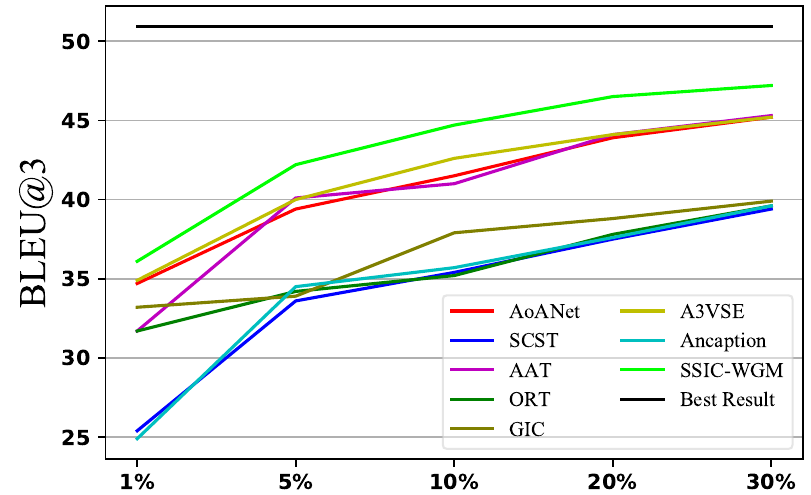}\\
			\mbox{ \;\;\;\; ({\it c1}) {BLEU$@$3 (XE)}}
		\end{minipage}
		\begin{minipage}[h]{43mm}
			\centering
			\includegraphics[width=43mm]{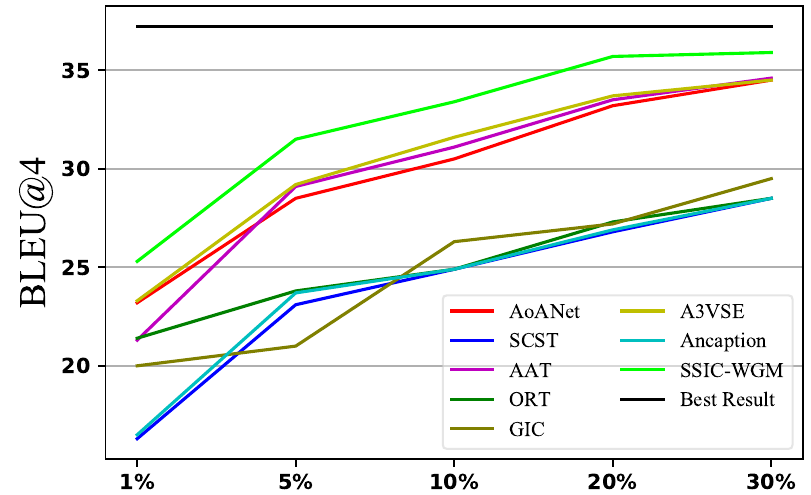}\\
			\mbox{ \;\;\;\; ({\it d1}) {BLEU$@$4 (XE)}}
		\end{minipage}\\
		\begin{minipage}[h]{43mm}
			\centering
			\includegraphics[width=43mm]{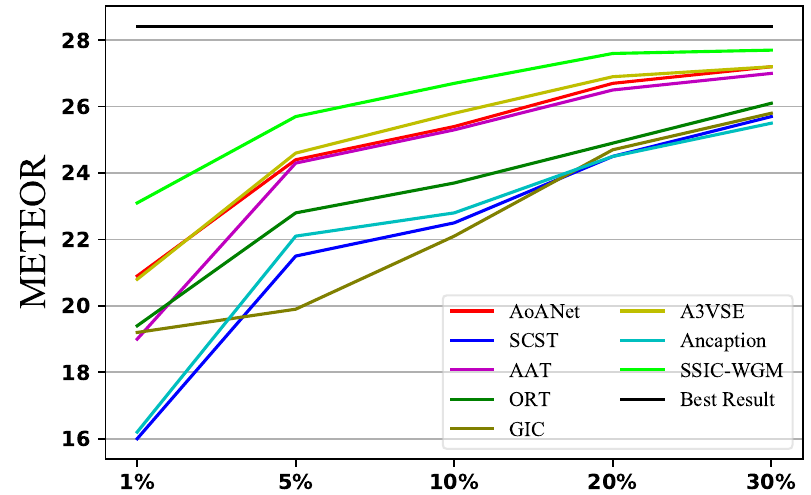}\\
			\mbox{ \;\;\;\; ({\it e1}) {METEOR (XE)}}
		\end{minipage}
		\begin{minipage}[h]{43mm}
			\centering
			\includegraphics[width=43mm]{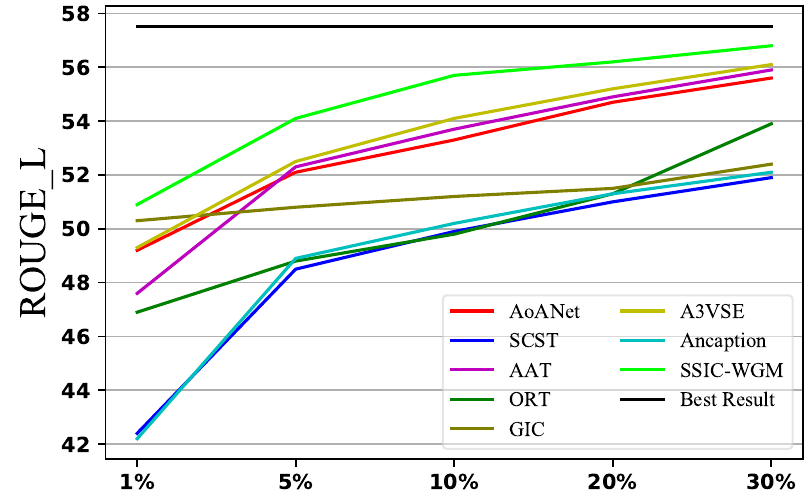}\\
			\mbox{ \;\;\;\; ({\it f1}) {ROUGE-L (XE)}}
		\end{minipage}
		\begin{minipage}[h]{43mm}
			\centering
			\includegraphics[width=43mm]{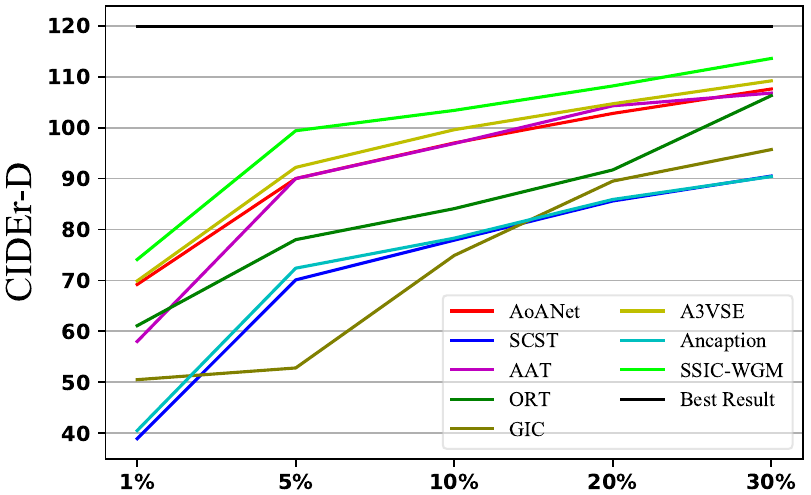}\\
			\mbox{ \;\;\;\; ({\it g1}) {CIDEr-D (XE)}}
		\end{minipage}
		\begin{minipage}[h]{43mm}
			\centering
			\includegraphics[width=43mm]{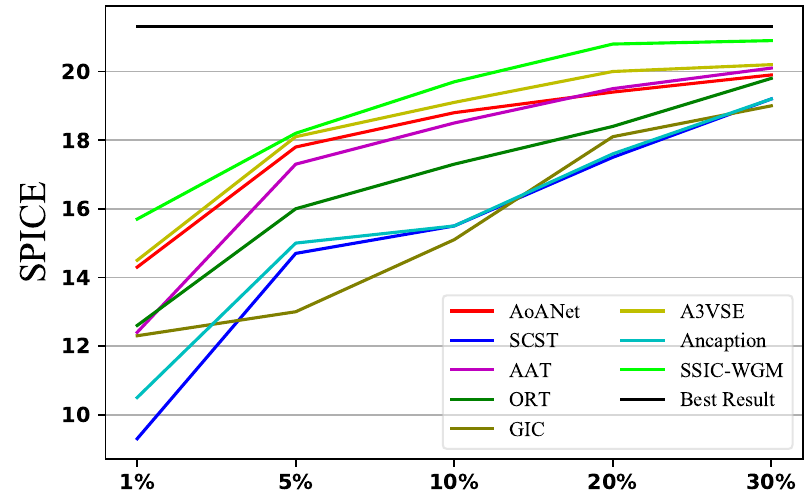}\\
			\mbox{ \;\;\;\; ({\it h1}) {SPICE (XE)}}
		\end{minipage}
		\begin{minipage}[h]{43mm}
			\centering
			\includegraphics[width=43mm]{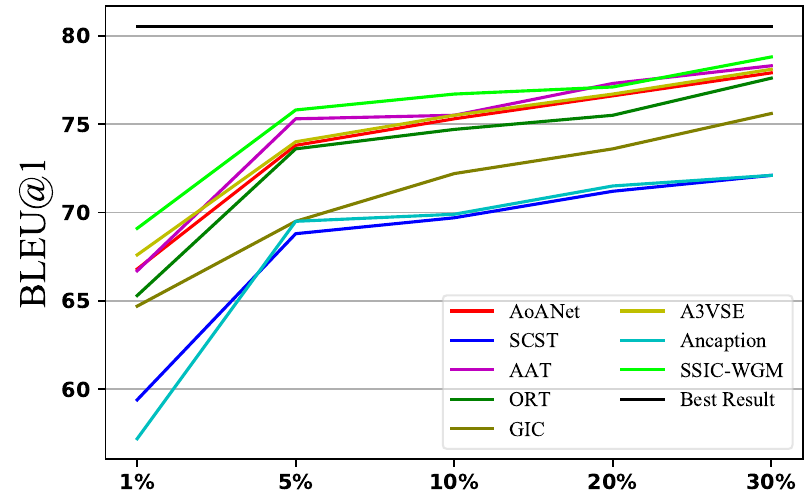}\\
			\mbox{ \;\;\;\; ({\it a2}) {BLEU$@$1 (RL)}}
		\end{minipage}
		\begin{minipage}[h]{43mm}
			\centering
			\includegraphics[width=43mm]{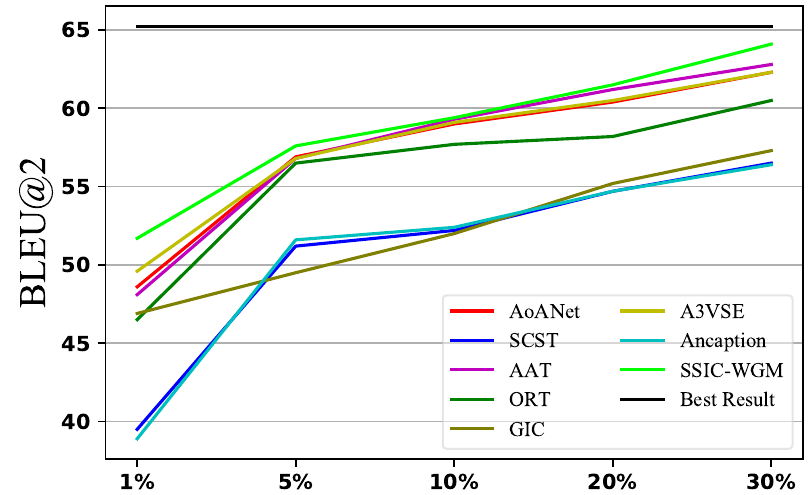}\\
			\mbox{ \;\;\;\; ({\it b2}) {BLEU$@$2 (RL)}}
		\end{minipage}
		\begin{minipage}[h]{43mm}
			\centering
			\includegraphics[width=43mm]{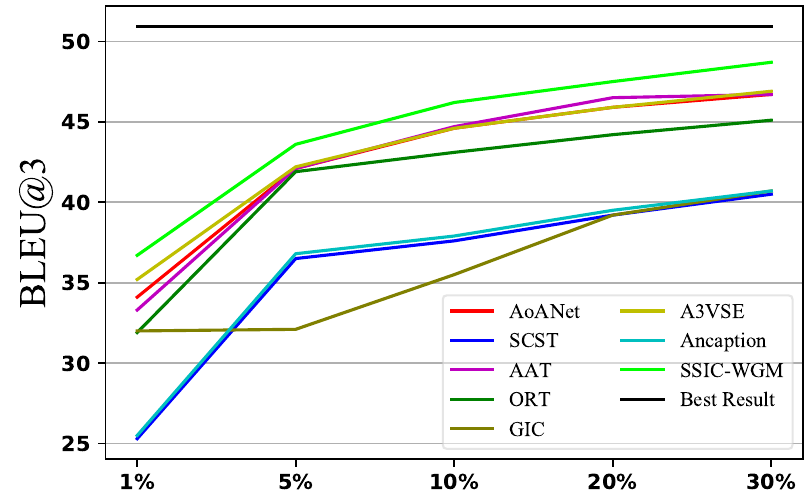}\\
			\mbox{ \;\;\;\; ({\it c2}) {BLEU$@$3 (RL)}}
		\end{minipage}
		\begin{minipage}[h]{43mm}
			\centering
			\includegraphics[width=43mm]{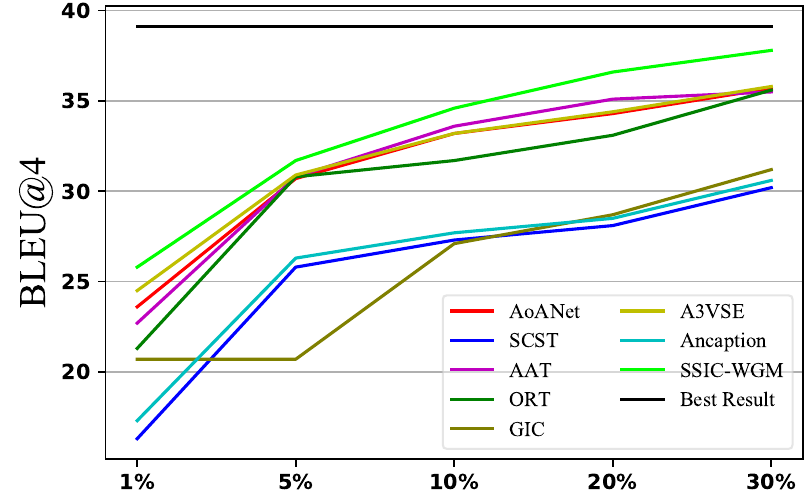}\\
			\mbox{ \;\;\;\; ({\it d2}) {BLEU$@$4 (RL)}}
		\end{minipage}\\
		\begin{minipage}[h]{43mm}
			\centering
			\includegraphics[width=43mm]{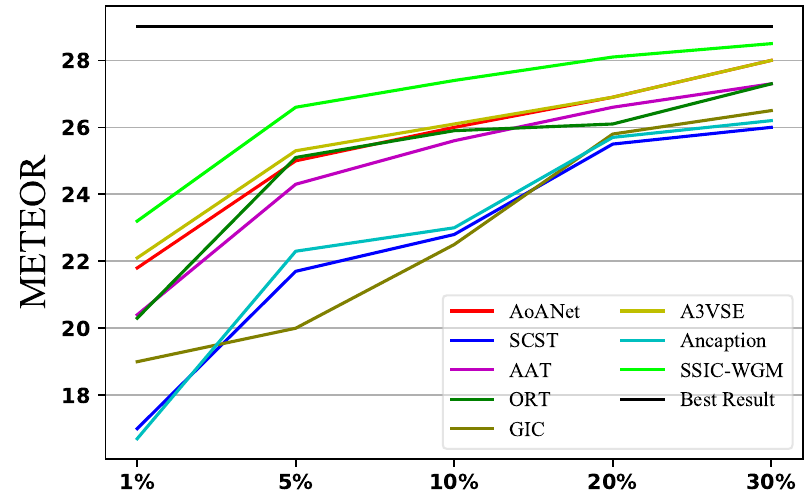}\\
			\mbox{ \;\;\;\; ({\it e2}) {METEOR (RL)}}
		\end{minipage}
		\begin{minipage}[h]{43mm}
			\centering
			\includegraphics[width=43mm]{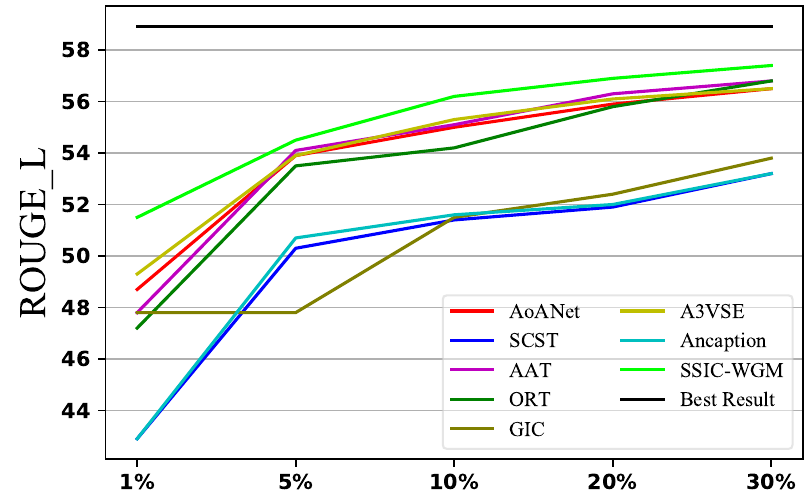}\\
			\mbox{ \;\;\;\; ({\it f2}) {ROUGE-L (RL)}}
		\end{minipage}
		\begin{minipage}[h]{43mm}
			\centering
			\includegraphics[width=43mm]{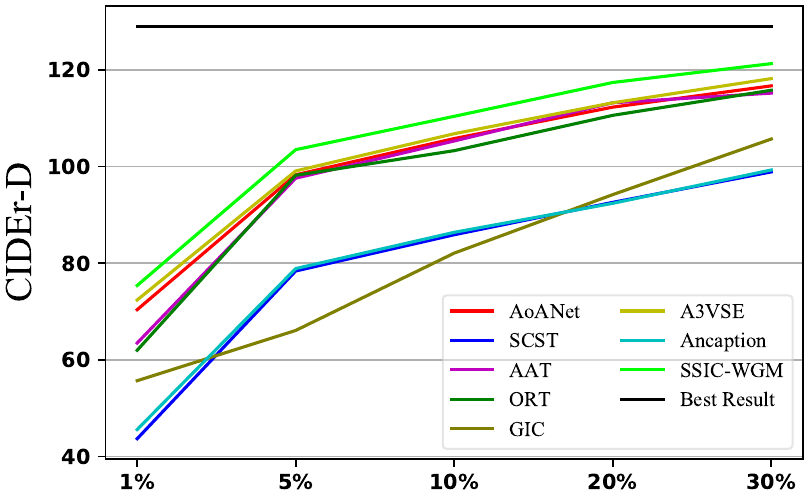}\\
			\mbox{ \;\;\;\; ({\it g2}) {CIDEr-D (RL)}}
		\end{minipage}
		\begin{minipage}[h]{43mm}
			\centering
			\includegraphics[width=43mm]{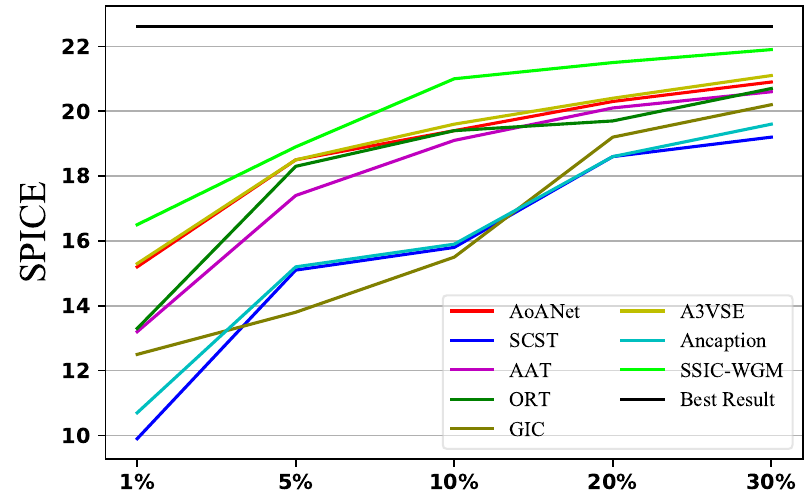}\\
			\mbox{ \;\;\;\; ({\it h2}) {SPICE (RL)}}
		\end{minipage}
	\end{center}
	\caption{Relationship between captioning performance with different ratio of supervised data, XE denotes the results of cross-entropy loss and RL represents the results of CIDEr-D Score Optimization.}\label{fig:f1}
\end{figure*}

\begin{figure}[!htb]
	\begin{center}
		\begin{minipage}[h]{80mm}
			\centering
			\includegraphics[width=80mm]{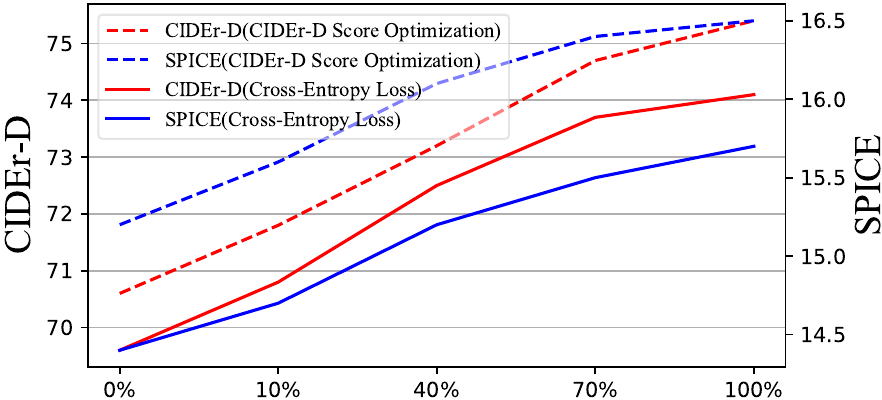}\\
		\end{minipage}
	\end{center}
	\caption{Relationship between captioning performance with different ratio of unsupervised data (CIDEr-D Score Optimization).}\label{fig:f2}
\end{figure}

\begin{table}[htb]{
		\centering
		\caption{Performance of SSIC-WGM with different augmentation number on MS-COCO “Karpathy” test splits. }
		\label{tab:tab3}
		\begin{tabular}{@{}l@{}|c|c|c|c|c|c|c|c}
			\toprule
			\multirow{2}{*}{Methods} & \multicolumn{8}{c}{Cross-Entropy Loss}  \\
			\cmidrule(l){2-9}
			& B$@$1 & B$@$2 & B$@$3 & B$@$4 & M & R & C & S \\
			\midrule
			K=1 &68.7&50.9&36.0&25.0&23.1&50.7&73.9&15.6\\
			K=2 &\bf68.8&\bf51.3&\bf36.1&\bf25.3&\bf23.1&\bf50.9&\bf74.1&\bf15.7\\
			K=3 &68.5&50.8&35.9&24.8&22.9&50.5&73.5&15.4\\
			K=4 &68.2&50.2&35.5&24.7&22.6&50.2&73.2&15.3\\
			\midrule
			\multirow{2}{*}{Methods}  & \multicolumn{8}{c}{CIDEr-D Score Optimization} \\
			\cmidrule(l){2-9}
			& B$@$1 & B$@$2 & B$@$3 & B$@$4 & M & R & C & S \\
			\midrule
			K=1 &68.9&51.6&36.5&25.5&23.1&51.3&75.2&16.3\\
			K=2 &\bf69.1&\bf51.7&\bf36.7&\bf25.8&\bf23.2&\bf51.5&\bf75.4&\bf16.5\\
			K=3 &68.7&51.1&36.3&25.1&23.0&50.8&74.5&16.2\\
			K=4 &68.5&50.5&36.1&24.9&22.9&50.6&73.9&15.8\\
			\bottomrule
	\end{tabular}}
\end{table}

\subsection{Ablation Study}
Moreover, we compare SSIC-WGM against other ablated models with various settings: 1) AoANet+Inter, we combine the inter-modal consistency (i.e., the representation consistency between global representations of input images' scene graph and generated sentences' scene graph for undescribed images) with the original AoANet generation loss of described images to train the model; 2) AoANet+Intra, we combine the intra-modal consistency (i.e., the representation consistency among global representations of generated sentences' scene graph for undescribed images) with the original AoANet generation loss of described images to train the model; 4) AoANet+Inter+Intra (AoANet+ for simplicity), we combine the inter-modal and intra-modal consistency defined above with the original AoANet generation loss; 5) SSL, we replace the intra-modal consistency with traditional semi-supervised methods (i.e., embedding consistency of generated sentences for undescribed images); 4) Strong+, we replace the weak augmentation with strong augmentation for SSIC-WGM; 6) GCN+, we conduct the node embedding as Eq. \ref{eq:e3} with hyper-parameters for SSIC-WGM; 7) w/o Inter, SSIC-WGM only retains the intra-modal consistency loss; 8) w/o Intra, SSIC-WGM only retains the inter-modal consistency. Note that the AoANet+Inter and AoANet+Intra adopt a naive aggregation for calculating the global representations of scene graphs (i.e., average all node embedding), and compare their substructures by calculating the Euclidean distance. Thereby we can verify the effectiveness of Wasserstein Distance (the difference between AoANet+Inter and w/o Intra is the calculation of graph distance). 

The bottom halves of Table \ref{tab:tab1} and Table \ref{tab:tab4} present the results: 1) AoANet+Inter and AoANet+Intra achieve better performance than AoANet, which indicates that both the intra-modal consistency and inter-modal consistency can improve the generator learning, and AoANet+Inter performs better than AoANet+Intra on most metrics, which indicates that inter-modal consistency is more significant, and the effect is also validated by w/o Inter and w/o Intra; 2) SSL performs worse than embedding based variants, which verifies that traditional semi-supervised techniques considering pseudo labeling are not as good as semantic and structure measurement considering raw image as pseudo supervision; 3) AoANet+Intra performs better than the SSL on most criteria (especially the CIDEr-D and SPICE score) reveal that the scene graph can better measure the similarity between the input image and the generated sentence; 4) Strong+ performs worse than the SSIC-WGM, which reveals that strong augmentation actually impact the generation of sentences; 5) GCN+ performs worse than the SSIC-WGM, for the reason that the simple scene graph structure is not conducive to the learning of weight parameters; 6) Both the w/o Inter and w/o Intra can improve the captioning performance on most criteria, especially the important criteria, i.e., CIDEr-D and SPICE, which indicates that each term can provide effective supervision to ensure the quality of generated sentences. On the other hand,  w/o Inter and w/o Intra perform better than the AoANet+Inter and AoANet+Intra, since the graph wasserstein distance considering region embedding is better than the distance measurement using naive averaged embedding;  7) SSIC-WGM performs competitive with SSIC-WGM+, for the reason that we have constrained the consistency of generated sentences' scene graphs, and at the same time constrained the consistency of images' scene graphs and generated sentences' scene graphs, so potentially constrained the consistency of images' scene graphs; 8) SSIC-WGM performs better than the AoANet+Inter+Intra, which validates the effectiveness of Wasserstein distance on graphs; and 9) SSIC-WGM and SSIC-WGM+ achieve the best scores on all metrics, which indicates that it is better to combine the inter-modal and intra-modal information. The limited improvement of some criteria is due to the lack of supervised data. Weighing the running efficiency and performance, the following comparison experiments all use the SSIC-WGM.

We add more experiments to validate the importance of each proposed components. In detail, we tune the coefficients separately for w/o Inter and w/o Intra in $\{0.01,0.1,1,10\}$. Figure \ref{fig:f10} records the results on MS-COCO dataset with cross entropy training, we find that the performance of intra-modal consistency increases with the increase of balance parameter, while the performance of inter-modal consistency gets better first, and then decreases when the balance parameter is larger than 1. Meanwhile, the performance of inter-modal consistency is better under the same parameter setting. The phenomena shows that inter-modal consistency is more important than intra-modal consistency, for the reason that inter-modal consistency can use the raw image information to directly constrain the generated sentences, whereas the generated sentences may introduce extra noises for intra-modal consistency. However, excessive attention to inter-modal consistency will affect the learning of supervised data, thereby reducing performance.

\subsection{Influence of the Supervised and Unsupervised Images}
To explore the influence of supervised and unsupervised data, we tune the ratio of supervised or unsupervised data to compare the performance. Figure \ref{fig:f1}  records the results with different ratio of supervised data under two metrics on MS-COCO dataset. We can find that with the percentage of supervised data increase, the performance of SSIC-WGM improves faster than other state-of-the-art methods, besides, note that SSIC-WGM performs competitive with supervised approach using all data when 30$\%$ training data are supervised, which validates SSIC-WGM can reasonably utilize the undescribed images to improve the learning of mapping function. Furthermore, Figure \ref{fig:f2} records the influence of unsupervised data on MS-COCO dataset, i.e., we fix the supervised ratio to 1$\%$, and tune the ratio of unsupervised data in $\{10\%, 40\%, 70\%, 100\%\}$. Considering the page limitation, we only give the results of two important criteria (i.e., SPICE and CIDEr-D). The results reveal that with the percentage of unsupervised data increases, the performance of SSIC-WGM also improves. This indicates that SSIC-WGM can make full use of undescribed images for positive training.

\subsection{Influence of the Augmentation Number}
To explore the influence of augmentation number, i.e., $K$, we tune the $K$ in $\{1,2,3,4\}$ and record the results on MS-COCO dataset in Table \ref{tab:tab3}. The results reveal that the SSIC-WGM achieves best performance with $K = 2$, for the reason that additional inconsistent noises between image and sentence may be introduced with the increasing of the number of augmentations.

\begin{table}[htb]{
		\centering
		\caption{Performance of SSIC-WGM with different caption model on MS-COCO “Karpathy” test split. ``+'' denotes adding the SSIC-WGM framework.}
		\label{tab:tab2}
		\begin{tabular}{@{}l@{}|@{}c|@{}c|c|c|c|c|c|c}
			\toprule
			\multirow{2}{*}{Methods} & \multicolumn{8}{c}{Cross-Entropy Loss}  \\
			\cmidrule(l){2-9}
			& B$@$1 & B$@$2 & B$@$3 & B$@$4 & M & R & C & S \\
			\midrule
			SCST  &56.8&38.6&25.4&16.3&16.0 &42.4&38.9&9.3\\
			GIC &63.0&46.8&33.2 &20.0&19.2&50.3&50.5&12.3\\
			Ancaption &56.5&38.1&24.9 &16.5&16.2&42.2&40.5&10.5\\
			\midrule
			SCST+ &\bf63.2&\bf45.7&\bf31.5&\bf21.5&\bf19.3&\bf45.4&\bf48.0&\bf10.2\\
			GIC+ &\bf66.9&\bf47.5&\bf34.8&\bf21.6&\bf19.6&\bf51.2&\bf57.8&\bf13.5\\
			Ancaption+ &\bf63.5&\bf45.8&\bf31.7&\bf21.6&\bf19.3&\bf45.6&\bf48.5&\bf10.9\\
			\midrule
			\multirow{2}{*}{Methods}  & \multicolumn{8}{c}{CIDEr-D Score Optimization} \\
			\cmidrule(l){2-9}
			& B$@$1 & B$@$2 & B$@$3 & B$@$4 & M & R & C & S \\
			\midrule
			SCST &59.4 &39.5&25.3 &16.3&17.0&42.9&43.7&9.9\\
			GIC &64.7 &46.9&32.0&20.7&19.0 &47.8&55.7&12.5\\
			Ancaption &57.2 &38.9&25.5&17.3&16.7 &42.9&45.6&10.7\\
			\midrule
			SCST+ &\bf66.2&\bf48.0&\bf33.7&\bf22.4&\bf20.2&\bf46.2&\bf48.5&\bf10.5\\
			GIC+ &\bf67.1&\bf48.5&\bf35.2&\bf22.6&\bf20.5&\bf49.7&\bf59.1&\bf13.9\\
			Ancaption+ &\bf66.3&\bf48.2&\bf33.7&\bf22.7&\bf20.3&\bf46.5&\bf49.2&\bf11.2\\
			\bottomrule
	\end{tabular}}
\end{table}

\begin{figure*}[t]
	\centering
	\includegraphics[width=180mm]{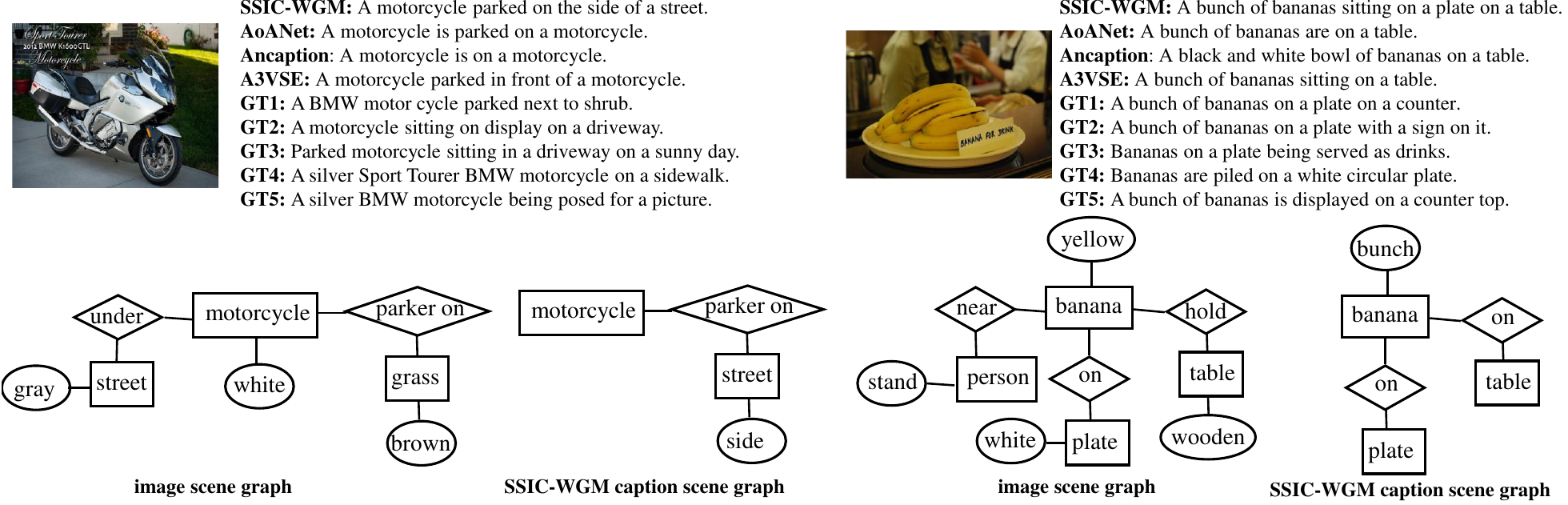}\\
	\caption{Examples of captions generated by SSIC-WGM and baseline models as well as the corresponding ground truths.}\label{fig:f3}
\end{figure*}

\begin{figure*}[!htb]
	\centering
	\includegraphics[width=185mm]{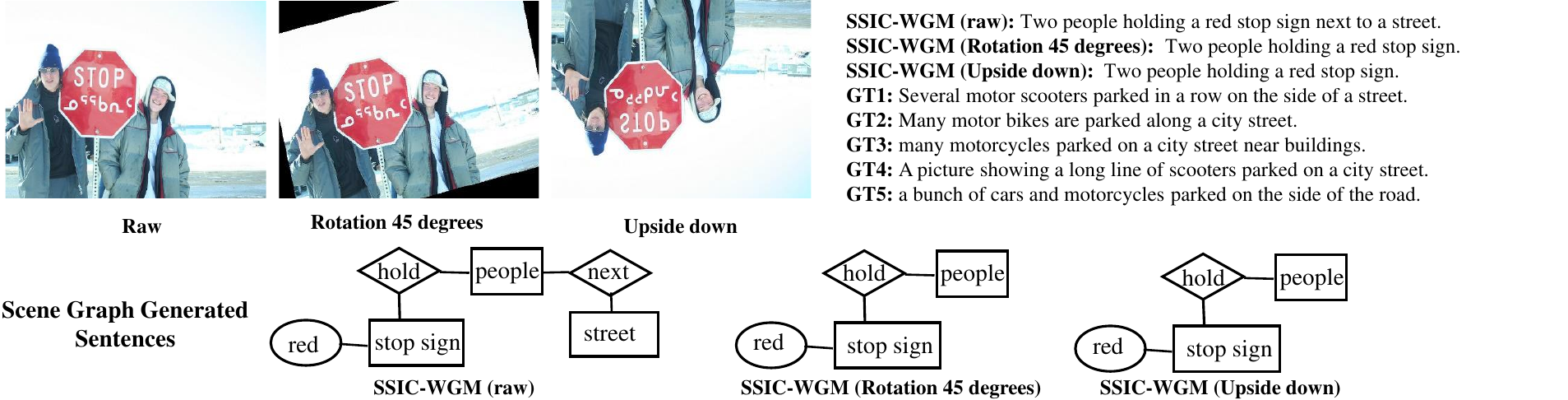}\\
	\caption{(Best viewed in color) Examples of captions generated by augmented images.}\label{fig:f4}
\end{figure*}

\begin{figure}[!htb]
	\begin{center}
		\begin{minipage}[h]{43mm}
			\centering
			\includegraphics[width=43mm]{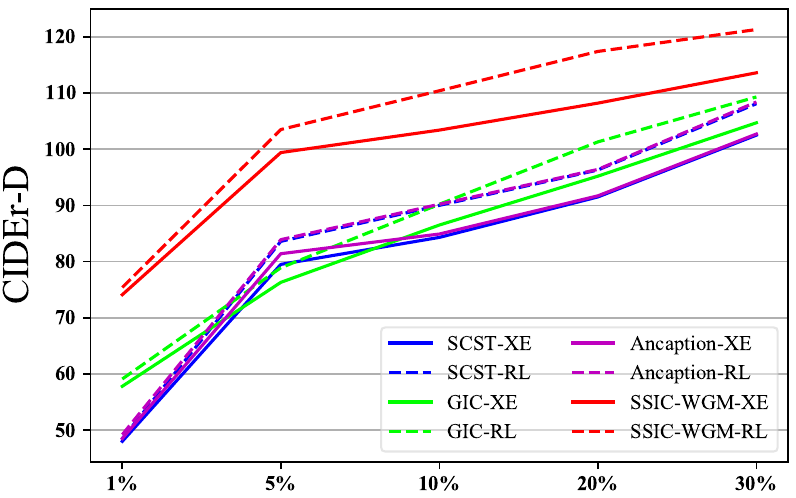}\\
			\mbox{ \;\;\;\; ({\it a}) {CIDEr-D}}
		\end{minipage}
		\begin{minipage}[h]{43mm}
			\centering
			\includegraphics[width=43mm]{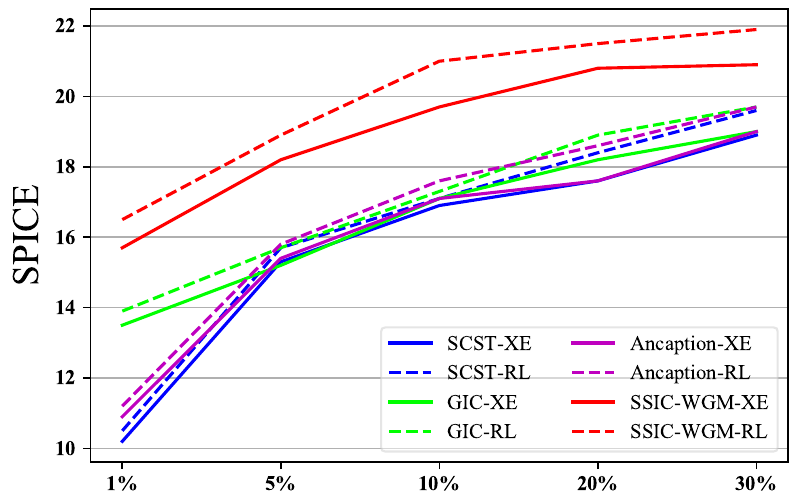}\\
			\mbox{ \;\;\;\; ({\it b}) {SPICE}}
		\end{minipage}
	\end{center}
	\caption{Relationship between captioning performance with different ratio of supervised data on MS-COCO dataset (CIDEr-D Score Optimization).}\label{fig:f7}
\end{figure}

\subsection{SSIC-WGM with Different Captioning Model}
To explore the generality of SSIC-WGM, we conduct more experiments by incorporating SSIC-WGM with different SOTA supervised captioning approaches, i.e., SCST (encoder-decoder based model), GIC and Ancaption (attention based models). The results of MS-COCO dataset are recorded in Table \ref{tab:tab2}. We find that all the methods, i.e., SCST, GIC, and Ancaption, have improved the performance after combing the SSIC-WGM framework. This phenomena validates that SSIC-WGM can well combine the undescribed image information for existing supervised captioning models.

To analyze the limitation increase of reinforcement learning optimization after using the proposed SSIC-WGM framework, we have added more experimental analysis and ablation experiments. At present, state-of-the-art models usually use reinforcement learning optimization to further fine-tune the model trained with cross-entropy loss through supervised data. In Table \ref{tab:tab2}, the performance improvement of CIDEr-D score optimization is limited after adding SSIC-WGM framework. The reason is that training data only has 1$\%$ described images, and the rest are all undescribed images. The supervised caption models are only trained with the described images, so the improvement of CIDEr-D score optimization is large. However, the SSIC-WGM framework uses all data (including described and undescribed images) for training. Therefore, the optimization process is biased to optimize the unsupervised loss, so compare to the supervised models, the performance of most criteria improves a lot, but the improvements of fine-tuning (i.e., CIDEr-D Score Optimization) are limited. Furthermore, we conduct more experiments to verify the effectiveness of the ratio of described images. In detail, we tune the ratio of described images in $\{1\%,5\%,10\%,20\%,30\%\}$, and record the results of MS-COCO dataset in Figure 7. We find that, with the increase of described data, the performance improvements of CIDEr-D score optimization also gradually increase, i.e., for SCST, Ancaption, and SSIC-WGM approaches, the improvements of CIDEr-D score optimization after using the proposed SSIC-WGM framework have increased by at least 5$\%$ when the described data accounted for 10$\%$, and the improvement of CIDEr-D score optimization after using the proposed SSIC-WGM framework has increased by 6.1$\%$ when the described data accounted for 20$\%$ for GIC.

%

\subsection{Visualization and Analysis}
Figure~\ref{fig:f3} shows a few examples with captions generated by our SSIC-WGM, three baselines, i.e., A3VSE, Ancaption and AoANet, and the human-annotated ground truths (i.e., $GT1-GT5$). From these examples, we find that the SSIC-WGM can generate accurate captions with high quality, while generated captions of baseline models lack the language structure and may be inaccurate for the image content. Figure~\ref{fig:f4} shows examples of augmented images and corresponding generated captions. From these examples, we find that the generated captions basically have similar semantic information, which can help the inter-modal and intra-modal consistencies for the undescribed images.

\section{Conclusion}\label{sec:s3}
Considering the cost of labor and data acquisition, semi-supervised image captioning methods have received widespread attentions and researches. Compared with traditional category labels, sentences are more complicated as label information because of complex context and semantic information. Therefore, the key challenge of semi-supervised image captioning is to robustly measure the sentences generated by undescribed images. However, existing self-supervision based or graph representation based approaches cannot effectively solve this problem, for the reason that generated sentences as pseudo labels have extra noises and the global graph embedding distance cannot effectively consider the local consistency. To address these problems, we introduce the Semi-Supervised Image Captioning considering Wasserstein Graph Matching (SSIC-WGM). SSIC-WGM can constrain the generated sentences from two aspects: 1) Inter-modal common space measurement. SSIC-WGM maps images and sentences to the high-order semantic spaces and constructs corresponding scene graphs for subsequent measurement, rather than uses generated sentence for self-supervision. 2) Wasserstein graph matching. SSIC-WGM employs the wasserstein distance to better measure the similarity between region embedding of different graphs, so as to avoid the shortcomings of global graph measurement. In result, SSIC-WGM learned an efficient mapping function under the semi-supervised scenario, and can be easily extended to any SOTA supervised approaches for handling undescribed images.



%

%
%

\ifCLASSOPTIONcompsoc
  \section*{Acknowledgments}
\else
  \section*{Acknowledgment}
\fi

This research was supported by NSFC (62006118,61836013,61906092,91746301), Natural Science Foundation of Jiangsu Province of China under Grant (BK20200460,BK20190441), CAAI-Huawei MindSpore Open Fund (CAAIXSJLJJ-2021-014B). Thank you mom for your support and wish you good health.

\ifCLASSOPTIONcaptionsoff
  \newpage
\fi

\bibliographystyle{IEEEtran}{
	\bibliography{SSIC-clean}}

\begin{thebibliography}{10}
\providecommand{\url}[1]{#1}
\csname url@samestyle\endcsname
\providecommand{\newblock}{\relax}
\providecommand{\bibinfo}[2]{#2}
\providecommand{\BIBentrySTDinterwordspacing}{\spaceskip=0pt\relax}
\providecommand{\BIBentryALTinterwordstretchfactor}{4}
\providecommand{\BIBentryALTinterwordspacing}{\spaceskip=\fontdimen2\font plus
\BIBentryALTinterwordstretchfactor\fontdimen3\font minus \fontdimen4\font\relax}
\providecommand{\BIBforeignlanguage}[2]{{%
\expandafter\ifx\csname l@#1\endcsname\relax
\typeout{** WARNING: IEEEtran.bst: No hyphenation pattern has been}%
\typeout{** loaded for the language `#1'. Using the pattern for}%
\typeout{** the default language instead.}%
\else
\language=\csname l@#1\endcsname
\fi
#2}}
\providecommand{\BIBdecl}{\relax}
\BIBdecl

\bibitem{Hossain2018}
M.~Z. Hossain, F.~Sohel, M.~F. Shiratuddin, and H.~Laga, ``A comprehensive survey of deep learning for image captioning,'' \emph{CoRR}, vol. abs/1810.04020, 2018.

\bibitem{KarpathyL15}
A.~Karpathy and F.~Li, ``Deep visual-semantic alignments for generating image descriptions,'' in \emph{CVPR}, Boston, MA, 2015, pp. 3128--3137.

\bibitem{ZhangP20}
J.~Zhang and Y.~Peng, ``Video captioning with object-aware spatio-temporal correlation and aggregation,'' \emph{{IEEE} Trans. Image Process.}, vol.~29, pp. 6209--6222, 2020.

\bibitem{GuptaVJ12}
A.~Gupta, Y.~Verma, and C.~V. Jawahar, ``Choosing linguistics over vision to describe images,'' in \emph{Proceedings of the 26th {AAAI} Conference on Artificial Intelligence}, Toronto, Ontario, 2012.

\bibitem{OrdonezKB11}
V.~Ordonez, G.~Kulkarni, and T.~L. Berg, ``Im2text: Describing images using 1 million captioned photographs,'' in \emph{Advances in Neural Information Processing Systems 24}, Granada, Spain, 2011, pp. 1143--1151.

\bibitem{YangTDA11}
Y.~Yang, C.~L. Teo, H.~D. III, and Y.~Aloimonos, ``Corpus-guided sentence generation of natural images,'' in \emph{Proceedings of the 2011 Conference on Empirical Methods in Natural Language Processing}, Edinburgh, UK, 2011, pp. 444--454.

\bibitem{FarabetCNL13}
C.~Farabet, C.~Couprie, L.~Najman, and Y.~LeCun, ``Learning hierarchical features for scene labeling,'' \emph{{IEEE} Trans. Pattern Anal. Mach. Intell.}, vol.~35, no.~8, pp. 1915--1929, 2013.

\bibitem{IjjinaM16}
E.~P. Ijjina and C.~K. Mohan, ``Hybrid deep neural network model for human action recognition,'' \emph{Appl. Soft Comput.}, vol.~46, pp. 936--952, 2016.

\bibitem{CollobertW08}
R.~Collobert and J.~Weston, ``A unified architecture for natural language processing: deep neural networks with multitask learning,'' in \emph{Proceedings of the 25th International Conference Machine Learning}, Helsinki, Finland, 2008, pp. 160--167.

\bibitem{BahdanauCB14}
D.~Bahdanau, K.~Cho, and Y.~Bengio, ``Neural machine translation by jointly learning to align and translate,'' in \emph{Proceedings of the 3rd International Conference on Learning Representations}, San Diego, CA, 2015.

\bibitem{VaswaniSPUJGKP17}
A.~Vaswani, N.~Shazeer, N.~Parmar, J.~Uszkoreit, L.~Jones, A.~N. Gomez, L.~Kaiser, and I.~Polosukhin, ``Attention is all you need,'' in \emph{Advances in Neural Information Processing Systems 30}, Long Beach, CA, 2017, pp. 5998--6008.

\bibitem{MaoXYWY14a}
J.~Mao, W.~Xu, Y.~Yang, J.~Wang, and A.~L. Yuille, ``Deep captioning with multimodal recurrent neural networks (m-rnn),'' in \emph{Proceedings of the 3rd International Conference on Learning Representations}, San Diego, CA, 2015.

\bibitem{XuBKCCSZB15}
K.~Xu, J.~Ba, R.~Kiros, K.~Cho, A.~C. Courville, R.~Salakhutdinov, R.~S. Zemel, and Y.~Bengio, ``Show, attend and tell: Neural image caption generation with visual attention,'' in \emph{ICML}, Lille, France, 2015, pp. 2048--2057.

\bibitem{SammaniE19}
F.~Sammani and M.~Elsayed, ``Look and modify: Modification networks for image captioning,'' in \emph{BMVC}, Cardiff, UK, 2019, p.~75.

\bibitem{ChoMGBBSB14}
K.~Cho, B.~van Merrienboer, cCaglar Gulccehre, D.~Bahdanau, F.~Bougares, H.~Schwenk, and Y.~Bengio, ``Learning phrase representations using {RNN} encoder-decoder for statistical machine translation,'' in \emph{Proceedings of the 2014 Conference on Empirical Methods in Natural Language Processing}, Doha, Qatar, 2014, pp. 1724--1734.

\bibitem{YangYWCS16}
Z.~Yang, Y.~Yuan, Y.~Wu, W.~W. Cohen, and R.~Salakhutdinov, ``Review networks for caption generation,'' in \emph{NeurIPS}, Barcelona, Spain, 2016, pp. 2361--2369.

\bibitem{yang2022exploiting}
Y.~Yang, H.~Wei, H.~Zhu, D.~Yu, H.~Xiong, and J.~Yang, ``Exploiting cross-modal prediction and relation consistency for semisupervised image captioning,'' \emph{IEEE Transactions on Cybernetics}, 2022.

\bibitem{LuXPS17}
J.~Lu, C.~Xiong, D.~Parikh, and R.~Socher, ``Knowing when to look: Adaptive attention via a visual sentinel for image captioning,'' in \emph{CVPR}, Honolulu, HI, 2017, pp. 3242--3250.

\bibitem{HuangWCW19}
L.~Huang, W.~Wang, J.~Chen, and X.~Wei, ``Attention on attention for image captioning,'' in \emph{ICCV}, Seoul, Korea, 2019, pp. 4633--4642.

\bibitem{SammaniM20}
F.~Sammani and L.~Melas{-}Kyriazi, ``Show, edit and tell: {A} framework for editing image captions,'' in \emph{CVPR}, Seattle, WA, 2020, pp. 4807--4815.

\bibitem{Anderson2018}
P.~Anderson, X.~He, C.~Buehler, D.~Teney, M.~Johnson, S.~Gould, and L.~Zhang, ``Bottom-up and top-down attention for image captioning and visual question answering,'' in \emph{CVPR}, Salt Lake City, UT, 2018, pp. 6077--6086.

\bibitem{YaoPLM18}
T.~Yao, Y.~Pan, Y.~Li, and T.~Mei, ``Exploring visual relationship for image captioning,'' in \emph{Proceedings of the 15th European Conference Computer Vision}, Munich, Germany, 2018, pp. 711--727.

\bibitem{HashimotoGOL18}
T.~B. Hashimoto, K.~Guu, Y.~Oren, and P.~Liang, ``A retrieve-and-edit framework for predicting structured outputs,'' in \emph{NeurIPS}, Montreal, Canada, 2018, pp. 10\,073--10\,083.

\bibitem{1983Vision}
H.~B. Barlow, ``Vision: A computational investigation into the human representation and processing of visual information,'' \emph{Journal of Mathematical Psychology}, vol.~27, no.~1, pp. 107--110, 1983.

\bibitem{Feng00L19a}
Y.~Feng, L.~Ma, W.~Liu, and J.~Luo, ``Unsupervised image captioning,'' in \emph{CVPR}, Long Beach, CA, 2019, pp. 4125--4134.

\bibitem{GuJCZYW19}
J.~Gu, S.~R. Joty, J.~Cai, H.~Zhao, X.~Yang, and G.~Wang, ``Unpaired image captioning via scene graph alignments,'' in \emph{ICCV}, Seoul, Korea, 2019, pp. 10\,322--10\,331.

\bibitem{MithunPPR18}
N.~C. Mithun, R.~Panda, E.~E. Papalexakis, and A.~K. Roy{-}Chowdhury, ``Webly supervised joint embedding for cross-modal image-text retrieval,'' in \emph{ACMMM}, Seoul, Republic of Korea, 2018, pp. 1856--1864.

\bibitem{HuangKLCH19}
P.~Huang, G.~Kang, W.~Liu, X.~Chang, and A.~G. Hauptmann, ``Annotation efficient cross-modal retrieval with adversarial attentive alignment,'' in \emph{ACMMM}, Nice, France, 2019, pp. 1758--1767.

\bibitem{JohnsonGF18}
J.~Johnson, A.~Gupta, and L.~Fei{-}Fei, ``Image generation from scene graphs,'' in \emph{Proceedings of the 2018 {IEEE} Conference on Computer Vision and Pattern Recognition}, Salt Lake City, UT, 2018, pp. 1219--1228.

\bibitem{WangLZY18}
Y.~Wang, C.~Liu, X.~Zeng, and A.~L. Yuille, ``Scene graph parsing as dependency parsing,'' in \emph{Proceedings of the 2018 Conference of the North American Chapter of the Association for Computational Linguistics: Human Language Technologies}, New Orleans, Louisiana, 2018, pp. 397--407.

\bibitem{Ludger2009Optimal}
L.~Rüschendorf, ``Optimal transport. old and new,'' \emph{Jahresbericht der Deutschen Mathematiker-Vereinigung}, vol. 111, no.~2, pp. 18--21, 2009.

\bibitem{KolouriPTSR17}
S.~Kolouri, S.~R. Park, M.~Thorpe, D.~Slepcev, and G.~K. Rohde, ``Optimal mass transport: Signal processing and machine-learning applications,'' \emph{{IEEE} Signal Process. Mag.}, vol.~34, no.~4, pp. 43--59, 2017.

\bibitem{TogninalliGLRB19}
M.~Togninalli, M.~E. Ghisu, F.~Llinares{-}Lopez, B.~Rieck, and K.~M. Borgwardt, ``Wasserstein weisfeiler-lehman graph kernels,'' in \emph{Advances in Neural Information Processing Systems 32}, Vancouver, Canada, 2019, pp. 6436--6446.

\bibitem{FrenchMF18}
G.~French, M.~Mackiewicz, and M.~H. Fisher, ``Self-ensembling for visual domain adaptation,'' in \emph{ICLR}, Vancouver, Canada, 2018.

\bibitem{RennieMMRG17}
S.~J. Rennie, E.~Marcheret, Y.~Mroueh, J.~Ross, and V.~Goel, ``Self-critical sequence training for image captioning,'' in \emph{CVPR}, Honolulu, HI, USA, 2017, pp. 1179--1195.

\bibitem{ZhouWLHZ20}
Y.~Zhou, M.~Wang, D.~Liu, Z.~Hu, and H.~Zhang, ``More grounded image captioning by distilling image-text matching model,'' in \emph{CVPR}, 2020, pp. 4776--4785.

\bibitem{LinMBHPRDZ14}
T.~Lin, M.~Maire, S.~J. Belongie, J.~Hays, P.~Perona, D.~Ramanan, P.~Dollar, and C.~L. Zitnick, ``Microsoft coco: Common objects in context,'' in \emph{ECCV}, Zurich, Switzerland, 2014, pp. 740--755.

\bibitem{YoungLHH14}
P.~Young, A.~Lai, M.~Hodosh, and J.~Hockenmaier, ``From image descriptions to visual denotations: New similarity metrics for semantic inference over event descriptions,'' \emph{Trans. Assoc. Comput. Linguistics}, vol.~2, pp. 67--78, 2014.

\bibitem{KuznetsovaOBBC12}
P.~Kuznetsova, V.~Ordonez, A.~C. Berg, T.~L. Berg, and Y.~Choi, ``Collective generation of natural image descriptions,'' in \emph{Proceedings of the 50th Annual Meeting of the Association for Computational Linguistics}, Jeju Island, Korea, 2012, pp. 359--368.

\bibitem{LiKBBC11}
S.~Li, G.~Kulkarni, T.~L. Berg, A.~C. Berg, and Y.~Choi, ``Composing simple image descriptions using web-scale n-grams,'' in \emph{Proceedings of the 15th Conference on Computational Natural Language Learning}, Portland, Oregon, 2011, pp. 220--228.

\bibitem{YaoYLLZ10}
B.~Z. Yao, X.~Yang, L.~Lin, M.~W. Lee, and S.~C. Zhu, ``{I2T:} image parsing to text description,'' \emph{Proceedings of the {IEEE}}, vol.~98, no.~8, pp. 1485--1508, 2010.

\bibitem{VinyalsTBE15}
O.~Vinyals, A.~Toshev, S.~Bengio, and D.~Erhan, ``Show and tell: {A} neural image caption generator,'' in \emph{CVPR}, Boston, MA, 2015, pp. 3156--3164.

\bibitem{YouJWFL16}
Q.~You, H.~Jin, Z.~Wang, C.~Fang, and J.~Luo, ``Image captioning with semantic attention,'' in \emph{CVPR}, Las Vegas, NV, 2016, pp. 4651--4659.

\bibitem{TeneyLH17}
D.~Teney, L.~Liu, and A.~van~den Hengel, ``Graph-structured representations for visual question answering,'' in \emph{Proceedings of the {IEEE} Conference on Computer Vision and Pattern Recognition}, Honolulu, HI, 2017, pp. 3233--3241.

\bibitem{LiuZZW19}
D.~Liu, H.~Zhang, Z.~Zha, and F.~Wu, ``Learning to assemble neural module tree networks for visual grounding,'' in \emph{Proceedings of the 2019 {IEEE/CVF} International Conference on Computer Vision}, Seoul, Korea (South), 2019, pp. 4672--4681.

\bibitem{KipfW17}
T.~N. Kipf and M.~Welling, ``Semi-supervised classification with graph convolutional networks,'' in \emph{ICLR}, Toulon, France, 2017.

\bibitem{Becigneul2020}
G.~Becigneul, O.~Ganea, B.~Chen, R.~Barzilay, and T.~S. Jaakkola, ``Optimal transport graph neural networks,'' \emph{CoRR}, vol. abs/2006.04804, 2020.

\bibitem{GrandvaletB04}
Y.~Grandvalet and Y.~Bengio, ``Semi-supervised learning by entropy minimization,'' in \emph{NeurIPS}, British Columbia, Canada, 2004, pp. 529--536.

\bibitem{ArazoOAOM20}
E.~Arazo, D.~Ortego, P.~Albert, N.~E. O'Connor, and K.~McGuinness, ``Pseudo-labeling and confirmation bias in deep semi-supervised learning,'' in \emph{IJCNN}, Glasgow, United Kingdom, 2020, pp. 1--8.

\bibitem{xi2023robust}
W.~Xi, X.~Song, W.~Guo, and Y.~Yang, ``Robust semi-supervised learning for self-learning open-world classes,'' in \emph{2023 IEEE International Conference on Data Mining (ICDM)}.\hskip 1em plus 0.5em minus 0.4em\relax IEEE, 2023, pp. 658--667.

\bibitem{BachmanAP14}
P.~Bachman, O.~Alsharif, and D.~Precup, ``Learning with pseudo-ensembles,'' in \emph{NeurIPS}, Quebec, Canada, 2014, pp. 3365--3373.

\bibitem{yang2021s2osc}
Y.~Yang, H.~Wei, Z.-Q. Sun, G.-Y. Li, Y.~Zhou, H.~Xiong, and J.~Yang, ``S2osc: A holistic semi-supervised approach for open set classification,'' \emph{ACM Transactions on Knowledge Discovery from Data (TKDD)}, vol.~16, no.~2, pp. 1--27, 2021.

\bibitem{XieDHL020}
Q.~Xie, Z.~Dai, E.~H. Hovy, T.~Luong, and Q.~Le, ``Unsupervised data augmentation for consistency training,'' in \emph{NeurIPS}, virtual, 2020.

\bibitem{BerthelotCCKSZR20}
D.~Berthelot, N.~Carlini, E.~D. Cubuk, A.~Kurakin, K.~Sohn, H.~Zhang, and C.~Raffel, ``Remixmatch: Semi-supervised learning with distribution matching and augmentation anchoring,'' in \emph{ICLR}, Addis Ababa, Ethiopia, 2020.

\bibitem{Ghani02}
R.~Ghani, ``Combining labeled and unlabeled data for multiclass text categorization,'' in \emph{ICML}, C.~Sammut and A.~G. Hoffmann, Eds., New South Wales, Australia, 2002, pp. 187--194.

\bibitem{sindhwani2008rkhs}
V.~Sindhwani and D.~S. Rosenberg, ``An rkhs for multi-view learning and manifold co-regularization,'' in \emph{ICML}, 2008, pp. 976--983.

\bibitem{yang2019comprehensive}
Y.~Yang, K.-T. Wang, D.-C. Zhan, H.~Xiong, and Y.~Jiang, ``Comprehensive semi-supervised multi-modal learning.'' in \emph{IJCAI}, 2019, pp. 4092--4098.

\bibitem{HeZRS16}
K.~He, X.~Zhang, S.~Ren, and J.~Sun, ``Deep residual learning for image recognition,'' in \emph{Proceedings of the 2016 {IEEE} Conference on Computer Vision and Pattern Recognition}, Las Vegas, NV, 2016, pp. 770--778.

\bibitem{RenHG017}
S.~Ren, K.~He, R.~B. Girshick, and J.~Sun, ``Faster {R-CNN:} towards real-time object detection with region proposal networks,'' \emph{{IEEE} Trans. Pattern Anal. Mach. Intell.}, vol.~39, no.~6, pp. 1137--1149, 2017.

\bibitem{VedantamZP15}
R.~Vedantam, C.~L. Zitnick, and D.~Parikh, ``Cider: Consensus-based image description evaluation,'' in \emph{CVPR}, Boston, USA, 2015, pp. 4566--4575.

\bibitem{YangTZC19}
X.~Yang, K.~Tang, H.~Zhang, and J.~Cai, ``Auto-encoding scene graphs for image captioning,'' in \emph{CVPR}, Long Beach, CA, 2019, pp. 10\,685--10\,694.

\bibitem{VishwanathanSKB10}
S.~V.~N. Vishwanathan, N.~N. Schraudolph, R.~Kondor, and K.~M. Borgwardt, ``Graph kernels,'' \emph{J. Mach. Learn. Res.}, vol.~11, pp. 1201--1242, 2010.

\bibitem{Haussler1999}
P.~Haussler, ``Convolution kernels on discrete structures,'' \emph{Technical Report. University of California}, 1999.

\bibitem{RubnerTG00}
Y.~Rubner, C.~Tomasi, and L.~J. Guibas, ``The earth mover's distance as a metric for image retrieval,'' \emph{Int. J. Comput. Vis.}, vol.~40, no.~2, pp. 99--121, 2000.

\bibitem{Cubuk20}
E.~D. Cubuk, B.~Zoph, D.~Man{\'{e}}, V.~Vasudevan, and Q.~V. Le, ``Autoaugment: Learning augmentation policies from data,'' \emph{CoRR}, vol. abs/1805.09501, 2018.

\bibitem{XuNTLD021}
G.~Xu, S.~Niu, M.~Tan, Y.~Luo, Q.~Du, and Q.~Wu, ``Towards accurate text-based image captioning with content diversity exploration,'' in \emph{CVPR}, virtual, 2021, pp. 12\,637--12\,646.

\bibitem{HerdadeKBS19}
S.~Herdade, A.~Kappeler, K.~Boakye, and J.~Soares, ``Image captioning: Transforming objects into words,'' in \emph{NeurIPS}, Vancouver, Canada, 2019, pp. 11\,135--11\,145.

\bibitem{HuangWXC19}
L.~Huang, W.~Wang, Y.~Xia, and J.~Chen, ``Adaptively aligned image captioning via adaptive attention time,'' in \emph{NeurIPS}, Vancouver, Canada, 2019, pp. 8940--8949.

\bibitem{KarpathyF17}
A.~Karpathy and L.~Fei{-}Fei, ``Deep visual-semantic alignments for generating image descriptions,'' \emph{TPAMI}, vol.~39, no.~4, pp. 664--676, 2017.

\bibitem{Xu2021}
G.~Xu, S.~Niu, M.~Tan, Y.~Luo, Q.~Du, and Q.~Wu, ``Towards accurate text-based image captioning with content diversity exploration,'' \emph{CoRR}, vol. abs/2105.03236, 2021.

\bibitem{ZhangSLJZWHJ21}
X.~Zhang, X.~Sun, Y.~Luo, J.~Ji, Y.~Zhou, Y.~Wu, F.~Huang, and R.~Ji, ``Rstnet: Captioning with adaptive attention on visual and non-visual words,'' in \emph{CVPR}, virtual, 2021, pp. 15\,465--15\,474.

\bibitem{guo2020recurrent}
D.~Guo, Y.~Wang, P.~Song, and M.~Wang, ``Recurrent relational memory network for unsupervised image captioning,'' in \emph{IJCAI}, 2020, pp. 920--926.

\bibitem{honda-etal-2021-removing}
U.~H.~Y. Ushiku and A.~H. T. W.~Y. Matsumoto, ``Removing word-level spurious alignment between images and pseudo-captions in unsupervised image captioning,'' in \emph{EACL}, 2021, pp. 3692--3702.

\bibitem{laina2019towards}
L.~Iro, R.~Christian, and N.~Nassir, ``Towards unsupervised image captioning with shared multimodal embeddings,'' in \emph{ICCV}, 2019, pp. 7414--7424.

\bibitem{ben2021unpaired}
H.~Ben, Y.~Pan, Y.~Li, T.~Yao, R.~Hong, M.~Wang, and T.~Mei, ``Unpaired image captioning with semantic-constrained self-learning,'' \emph{Trans. Multimedia}, 2021.

\bibitem{ChenJZ21}
X.~Chen, M.~Jiang, and Q.~Zhao, ``Self-distillation for few-shot image captioning,'' in \emph{WACV}, Waikoloa, HI, 2021, pp. 545--555.

\bibitem{JainSJM021}
A.~Jain, P.~R. Samala, P.~Jyothi, D.~Mittal, and M.~K. Singh, ``Perturb, predict {\&} paraphrase: Semi-supervised learning using noisy student for image captioning,'' in \emph{IJCAI}, Virtual, 2021, pp. 758--764.

\bibitem{PapineniRWZ02}
K.~Papineni, S.~Roukos, T.~Ward, and W.~Zhu, ``Bleu: a method for automatic evaluation of machine translation,'' in \emph{ACL}, Philadelphia, USA, 2002, pp. 311--318.

\bibitem{BanerjeeL05}
S.~Banerjee and A.~Lavie, ``{METEOR:} an automatic metric for {MT} evaluation with improved correlation with human judgments,'' in \emph{IEEMMT}, AnnArbor, USA, 2005, pp. 65--72.

\bibitem{AndersonFJG16}
P.~Anderson, B.~Fernando, M.~Johnson, and S.~Gould, ``{SPICE:} semantic propositional image caption evaluation,'' in \emph{ECCV}, Amsterdam, The Netherlands, 2016, pp. 382--398.

\bibitem{GimenezM07a}
J.~Gim{\'{e}}nez and L.~M{\`{a}}rquez, ``Linguistic features for automatic evaluation of heterogenous {MT} systems,'' in \emph{ACL Workshop}, C.~Callison{-}Burch, P.~Koehn, C.~S. Fordyce, and C.~Monz, Eds., Prague, Czech Republic, 2007, pp. 256--264.

\end{thebibliography}

\vspace{-1.2cm}
\begin{IEEEbiography}[{\includegraphics[width=1in,height=1.25in,clip,keepaspectratio]{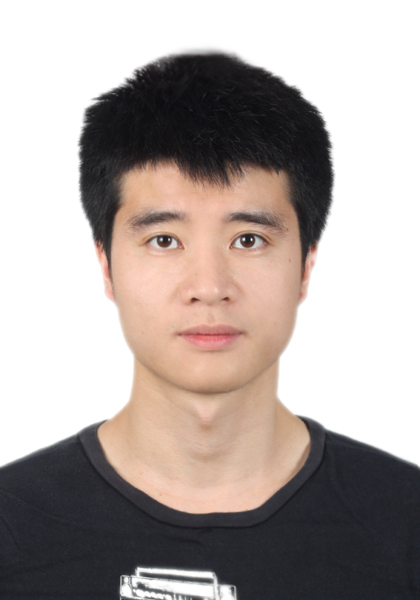}}]{Yang Yang}
	received the Ph.D. degree in computer science, Nanjing University, China in 2019. At the same year, he became a faculty member at Nanjing University of Science and Technology, China. He is currently a Professor with the School of Computer Science and Engineering. His research interests lie primarily in machine learning and data mining, including heterogeneous learning, model reuse, and incremental mining. He has published over 10 papers in leading international journal/conferences. He serves as PC in leading conferences such as IJCAI, AAAI, ICML, NIPS, etc.
\end{IEEEbiography}

\end{document}